\documentclass[10pt,twocolumn,letterpaper]{article}

\usepackage{iccv}              

\usepackage[pagebackref,breaklinks,colorlinks]{hyperref}
\usepackage{color}
\usepackage{caption, subcaption, multirow, overpic, textpos}
\usepackage{bbm}
\usepackage{titletoc}
\usepackage{graphicx}

\usepackage{colortbl}
\usepackage{tcolorbox}

\usepackage{multirow}
\usepackage{booktabs}
\usepackage{makecell}

\usepackage{amsmath}
\usepackage{amssymb}
\usepackage{mathtools}
\usepackage{amsthm}

\usepackage[capitalize,noabbrev]{cleveref}

\definecolor{iccvblue}{RGB}{0, 102, 204}
\hypersetup{
    colorlinks=true, 
    citecolor=iccvblue,
}

\usepackage{bm}
\usepackage{tocbasic}

\crefname{section}{Sec.}{Secs.}
\Crefname{section}{Section}{Sections}
\Crefname{table}{Table}{Tables}
\crefname{table}{Tab.}{Tabs.}
\crefname{figure}{Fig.}{Figs.}
\crefname{equation}{Eq.}{Eqs.}

\definecolor{Gray}{gray}{0.9}
\definecolor{LightCobaltBlue}{RGB}{143,170,220}
\definecolor{Amber}{RGB}{255,102,0}
\definecolor{BlackOlive}{RGB}{59,56,56}
\definecolor{AlloyOrange}{RGB}{197,90,17}
\definecolor{B'dazzledBlue}{RGB}{47,85,151}
\definecolor{DarkBlue}{RGB}{72, 116, 203}
\definecolor{DarkGreen}{RGB}{106, 169, 63}

\definecolor{baselinecolor}{gray}{.9}

\definecolor{iccvblue}{rgb}{0.21,0.49,0.74}

\def\paperID{305} 
\def\confName{ICCV}
\def\confYear{2025}

\newcommand{\tablestyle}[2]{\setlength{\tabcolsep}{#1}\renewcommand{\arraystretch}{#2}}
\newcommand{\baseline}[1]{\cellcolor{baselinecolor}{#1}}

\begin{document}

\title{Decoupled Diffusion Sparks 
Adaptive
Scene Generation}
\author{
{Yunsong Zhou}$^{1, 2\ast}$ \quad
{Naisheng Ye}$^{1,3\ast}$ \quad
{William Ljungbergh}$^{4}$ \quad
{Tianyu Li}$^{1}$ \quad
{Jiazhi Yang}$^{1}$ \quad \\
{Zetong Yang}$^{5}$ \quad
{Hongzi Zhu}$^{2}$ \quad
{Christoffer Petersson}$^{4}$ \quad
{Hongyang Li}$^{1}$ \\
[2mm]
$^1$~OpenDriveLab  
\quad
$^2$~Shanghai Jiao Tong University \quad
$^3$~Zhejiang University \quad \\
$^4$~Zenseact \quad 
$^5$~GAC R\&D Center
\\
[1mm]
\normalsize{
\url{https://opendrivelab.com/Nexus}
}}

\maketitle

{\let\thefootnote \relax \footnote{$^\ast$Equal contribution.}}

\begin{abstract}
   Controllable scene generation could reduce the cost of diverse data collection substantially for autonomous driving.
   Prior works formulate the traffic layout generation as predictive progress, either by denoising entire sequences at once or by iteratively predicting the next frame.
   However, full sequence denoising hinders online reaction, while the latter's short-sighted next-frame prediction lacks precise goal-state guidance.
   Further, the learned model struggles to generate complex or challenging scenarios due to a large number of safe and ordinal driving behaviors from open datasets.
   To overcome these, we introduce Nexus, a decoupled scene generation framework that improves reactivity and goal conditioning by simulating both ordinal and challenging scenarios from fine-grained tokens with independent noise states.
   At the core of the decoupled pipeline is the integration of a partial noise-masking training strategy and a noise-aware schedule that ensures timely environmental updates throughout the denoising process.
   To complement challenging scenario generation, we collect a dataset consisting of complex corner cases. It covers 540 hours of simulated data, including high-risk interactions such as cut-in, sudden braking, and collision.
   Nexus achieves superior generation realism while preserving reactivity and goal orientation, with a 40\% reduction in displacement error.
   We further demonstrate that Nexus improves closed-loop planning by 20\% through data augmentation and showcase its capability in safety-critical data generation.
   %

\end{abstract}

\begin{figure}[t]
\begin{center}
\centerline{\includegraphics[width=\linewidth]{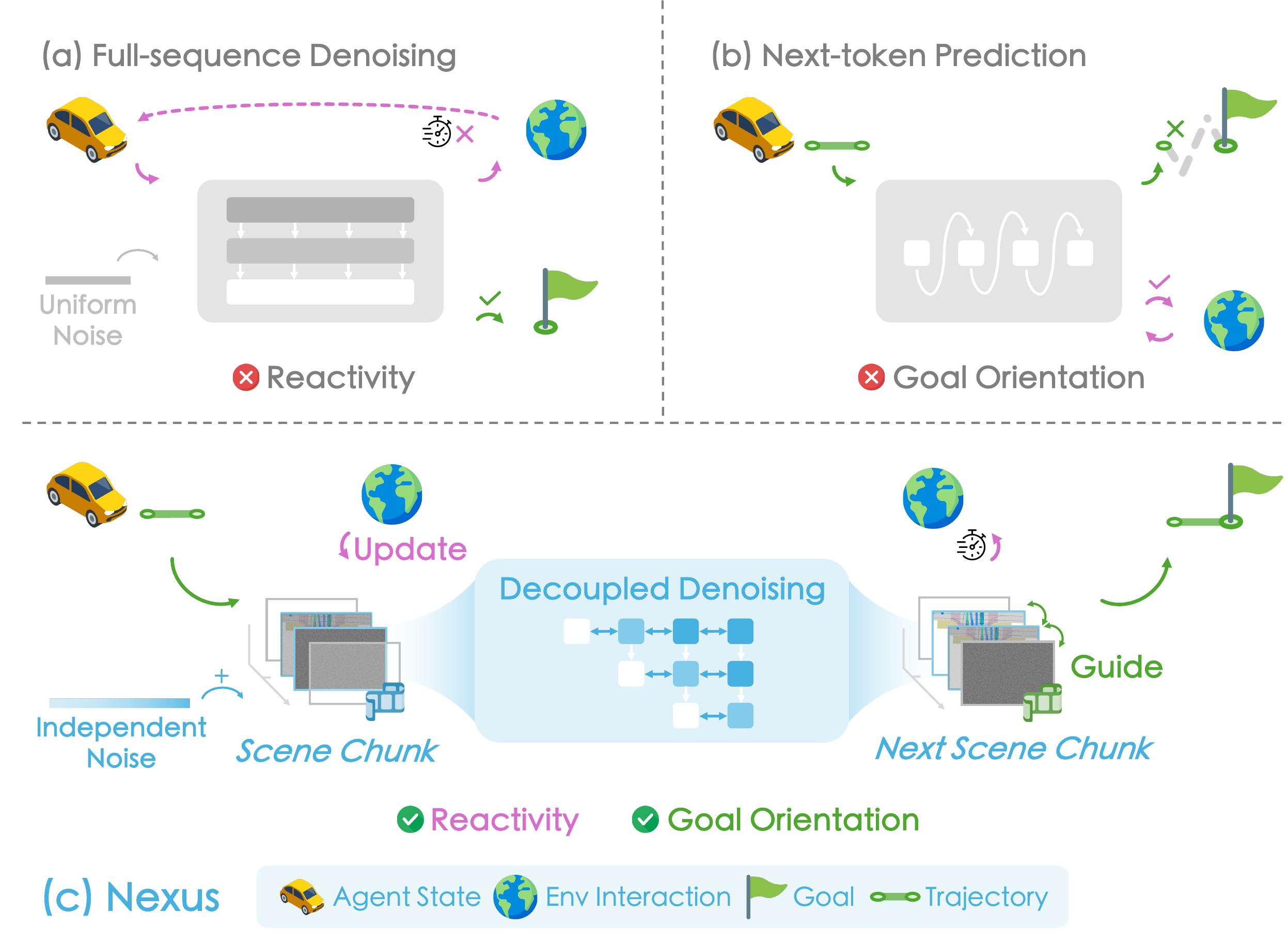}}
\caption{
\textbf{Nexus} is a noise-decoupled prediction pipeline designed for adaptive driving scene generation, ensuring both timely reaction and goal-directed control.
Unlike prior approaches that use \textbf{(a)} full-sequence denoising or \textbf{(b)} next-token prediction, \textbf{(c)} Nexus introduces \textit{independent} yet structured noise states, enabling more controlled and interactive scene generation.
%
%
It leverages low-noise goals to steer generation while incorporating environmental updates dynamically, which are captured in subsequent denoising.
%
}
\vspace{-20pt}
\label{fig:teaser}
\end{center}
\end{figure}

\section{Introduction}
\label{introduction}


Diversity is crucial for autonomous driving datasets, as data-driven solutions struggle with the scarcity of critical long-tail scenarios.
%
%
Due to the high cost of collecting rare long-tail driving data, high-fidelity world generators offer a cost-effective alternative for producing diverse scenarios with rare driving behaviors~\cite{ma2024unleashing}.
%
Besides delivering realistic visuals~\cite{gao2024vista,li2023drivingdiffusion,wen2023panacea}, crafting reasonable and diverse traffic layouts is vital for an adaptive generator to be applicable.
%
This drives two key requirements:
%
1) Reactivity, which incorporates environmental feedback to model interactions between agents and adjust scene evolution dynamically in response to real-time variations in driving decisions.
%
%
2) Goal orientation, which ensures controlled, non-stochastic scene generation guided by predefined future states, allowing the synthesis of realistic safety-critical scenarios with a well-defined outcome.
%

In this field, diffusion models~\cite{chi2023diffusion,yang2024diffusion} show promising results in generating realistic scene layouts conditioned on text prompts~\cite{tan2023language}, protocols~\cite{jiang2024scenediffuser}, and road maps~\cite{feng2023trafficgen}.
%
However, these models struggle to respond to real-time agent interactions due to their rigid full-sequence denoising process, which prevents immediate response to new environmental changes (\cref{fig:teaser} (a)).
%
Updates from interactions 
cannot affect the generation timely, 
forcing the model to discard previously generated future states and regenerate them entirely.
%
In addition, the rarity of critical scenarios in public datasets~\cite{caesar2021nuplan,sun2020scalability} limits their ability to generate diverse situations, as these datasets primarily capture routine driving behaviors and lack sufficient risky cases.
%
%
Alternatively, predictive transformers~\cite{hu2025solving,philion2024trajeglish,radford2019language}, which excel in responding to environments by continuously rolling out the next frames of the current scenario in \cref{fig:teaser} (b).
However, they lack awareness of the goal state, as future states are inaccessible to the model during causal generation, making precise control over safety-critical scenarios difficult.
%
Even with global contexts as guidance, precise controls like directing for a collision remain challenging.
%
As a result, existing approaches fail to simultaneously provide both real-time reactivity and goal-directed scene generation, limiting their use in high-fidelity world modeling.

To this end, we introduce \textbf{\texttt{Nexus}}, a decoupled predictive model that integrates independent \underline{n}ois\underline{e} a\underline{cross} diff\underline{us}ing steps, marrying the reactivity of predictive models with the goal-awareness of diffusion-based approaches.
%
As shown in \cref{fig:teaser} (c), Nexus integrates differentiated agent state with decoupled denoising, moving beyond full-sequence scenario generation by adaptively evolving scene chunks over time.
%
%
%
Each chunk, a localized subset of the scenario, encodes uncertainty using noise as a soft mask; low-noise regions guide generation, while high-noise tokens allow the reaction to new environmental changes.

%
%
%

Specific designs are proposed to achieve the functionality through the decoupled diffusion model.
%
%
For goal orientation, our noise-masking training strategy enables Nexus to reconstruct the original sequence from individually corrupted tokens.
%
This facilitates the flexible combination of low-noise goals and high-noise target scenarios during inference, free of adaptation for guided generation.
For reactivity, unlike slow autoregressive approaches, our noise-aware scheduling directly adapts token states, ensuring rapid response to environmental changes without unnecessary recomputation.
%
Changes are directly reflected by overwriting the corresponding token states, while the pipelined sampling distributes cost across frames and pops zero-noise tokens at each denoising step for interaction.
%
%
%
%
%
%
%

To foster a general goal orientation ability for rare or unseen corner cases, we construct the large corpus of safety-critical driving scenarios, \textbf{Nexus-Data}.
%
%
The simulator captures complex behaviors that seldom appear in real-world data through interactions with the physics engine.
We generate high-quality training data using the MetaDrive simulator~\cite{li2022metadrive}, where virtual traffic flows are synthesized via adversarial learning~\cite{zhang2023cat} and filtered through automated validity checks to ensure diverse and realistic driving interactions.
%
Nexus-Data comprises 540 hours of simulated driving, representing the largest-scale collection of challenging scenarios, including merging, cut-in, and collision.

We summarize our contributions as follows: 
1) We introduce Nexus, a decoupled diffusion model that enables adaptive scene generation by learning independent yet structured noise states, improving both goal conditioning and reactive scene updates.
%
%
2)
We propose noise-masking training, which allows Nexus to integrate goal conditioning with diverse scenario evolution seamlessly. Besides, our noise-aware scheduling mechanism ensures real-time responsiveness by selectively updating only relevant token states.
%
%
%
3) We construct Nexus-Data, a scaled dataset of high-risk driving scenes, enhancing the model's generalization to safety-critical cases. 
Building on this data and decoupled diffusion, Nexus surpasses Diffusion Policy~\cite{chi2023diffusion}, GUMP~\cite{hu2025solving}, and SceneDiffuser~\cite{jiang2024scenediffuser} in controllability, interactivity, and kinematics, reducing displacement error by 40\%.

\begin{figure}[t]
\begin{center}
\centerline{\includegraphics[width=\linewidth]{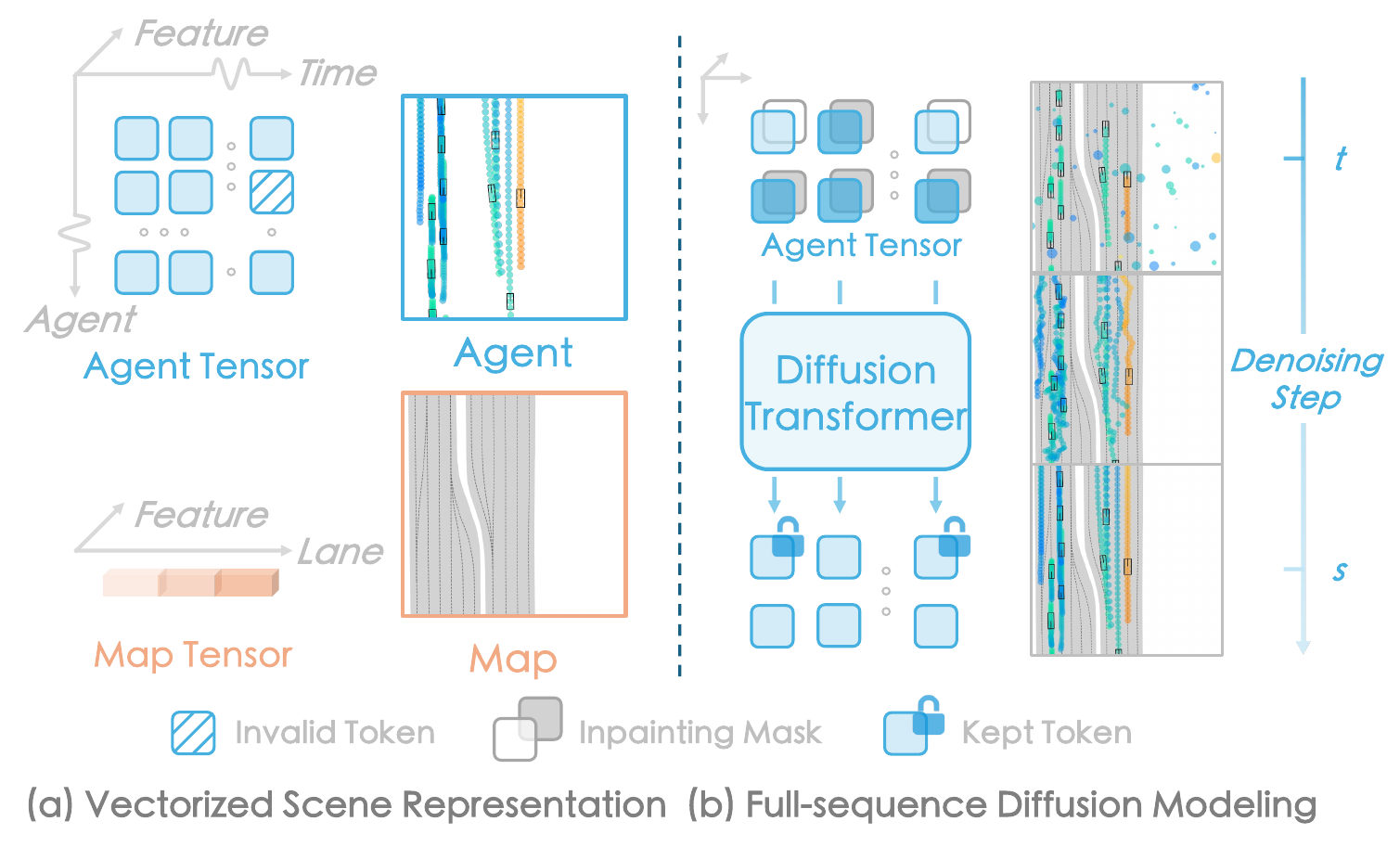}}
\caption{
\textbf{Preliminary on the scene generation.}
\textbf{(a)} Current methods encode scenes with tokens for agent and map attributes, formulating scene generation as generating future agent tensors from historical ones conditioned on a global map tensor.
\textbf{(b)} Diffusion models take the entire sequence as input, using hard masks to fix conditions and enable controllable generation via inpainting, yet fail in a timely reaction.
}
\vspace{-20pt}
\label{fig:preliminary}
\end{center}
\end{figure}

\begin{figure*}[t]
\begin{center}
\centerline{\includegraphics[width=\linewidth]{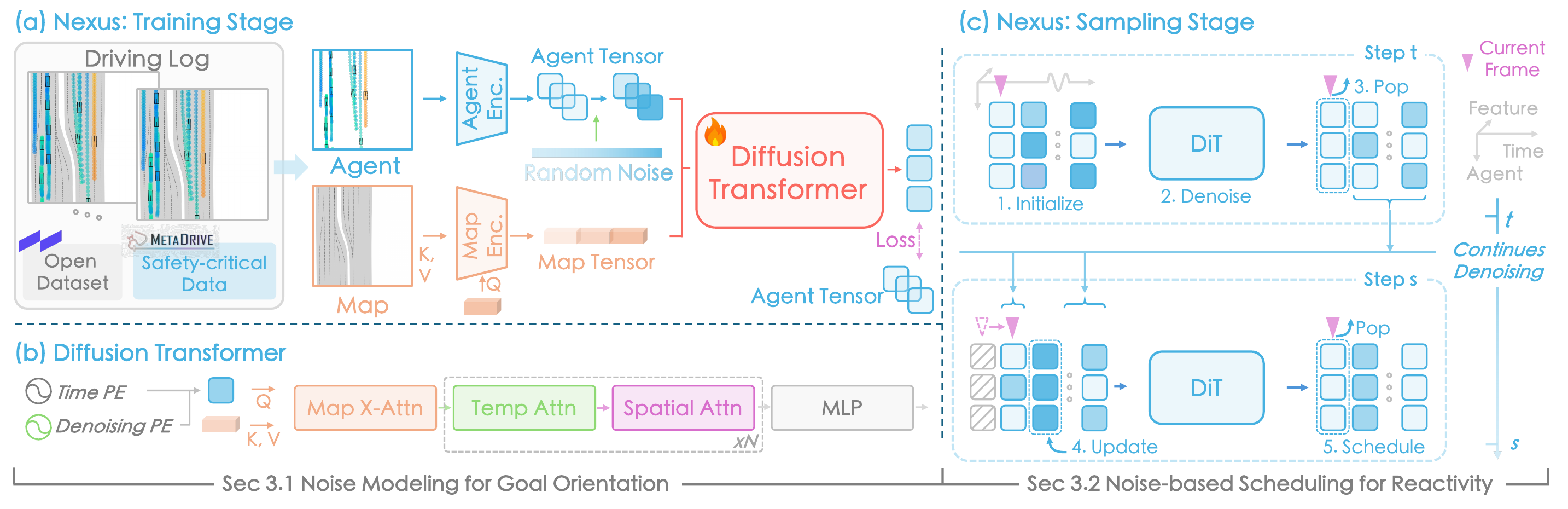}}
\caption{
\textbf{Framework of Nexus.}
\textbf{(a)} Nexus learns from realistic and safety-critical driving logs and encodes agents and maps separately before feeding them into a diffusion transformer. 
The model is trained to restore sequences from partially masked agent tokens guided by low-noise ones.
\textbf{(b)} Agent tokens are encoded with time and denoising steps, then interact with the maps and dynamics via attention.
\textbf{(c)} Tokens with varying noise are scheduled within a chunk for a timely reaction.
Each denoising step updates and pops zero-noise tokens, replacing them with next-frame tokens to iteratively generate the scene.
%
%
\vspace{-20pt}
}
\label{fig:pipeline}
\end{center}
\end{figure*}

\section{Preliminary}
\label{sec:preliminary}


Traffic layout simulation for autonomous driving requires structured representations of both agent behaviors and map features.
%
Recent works frame scene generation as a sequence modeling task, where driving scenarios are represented as structured tokenized states, enabling simultaneous prediction of all agent futures~\cite{hu2025solving,philion2024trajeglish,tan2024promptable}.


\noindent
\textbf{Vectorized representation of scene generation.}
As shown in \cref{fig:preliminary} (a), driving scenarios are encoded as structured token representations, consisting of an \textit{Agent Tensor} for dynamic entities and a \textit{Map Tensor} for static environment features.
We denote the agent tensor as $\mathbf{x} \in \mathbb{R}^{A\times \mathcal{T} \times D}$, where $A$ is the maximum number of agents, $\mathcal{T}$ is the number of physical timesteps, and $D$ is the dimension of agent attributes.
The attribute for each agent includes positional coordinates ($x$, $y$), heading ($sin_{\alpha}$, $cos_{\alpha}$), velocities ($v_x$, $v_y$), and dimensions ($l$, $w$).
A valid mask $\mathbf{m} \in \mathbb{B}^{A\times \mathcal{T}}$ is initialized to indicate which agents in the agent tensor $\mathbf{x}$ are valid at each timestep.
As for the map information, the map tensor $\mathbf{c}\in \mathbb{R}^{L\times N \times D^{'}}$ is used to represent the lanes' conditions, where $L$, $N$, and $D^{'}$ stand for the number of lanes, points per lane, and attributes (coordinates and types), respectively.
Based on the vectorized representation, sequential modeling of driving scenes can be expressed as generating the future scene tensor $\mathbf{x}\odot \mathbf{m}_{:,\tau:,:}$ given the current timestep $\tau<\mathcal{T}$, historical scene tensor $\mathbf{x}_{:,:\tau,:}$, and global map tensor $\mathbf{c}$.
To simplify the model's learning task, all feature channels are normalized with corresponding means and deviations before concatenating.


\noindent
\textbf{Full-sequence diffusion modeling.}
Diffusion transformers (DiTs)~\cite{peebles2023scalable} are a class of generative models that generate the agent tensor $\mathbf{x}$ by reversing a stochastic differential process
~\cite{jiang2024scenediffuser,yang2024diffusion}.
It can be implemented as stacked transformer blocks $\epsilon_{\theta}$.
Let $\mathbf{x}^0\in \mathcal{X}$ represent a latent feature from the distribution $p(\mathbf{x})$.
Training begins with an initial latent state $\mathbf{x}^0$, which undergoes progressive noise injection over timesteps $t \in (0,1]$ until reaching a Gaussian noise distribution at $\mathbf{x}^1$.
%
The model is optimized by minimizing the mean-square error (MSE):
\begin{align}
\mathbf{x}^t=\alpha_t \mathbf{x}^0+\sigma_t \epsilon, \epsilon & \sim \mathcal{N}(\mathbf{0},\mathbf{I}), \mathbf{x}^0 \sim p(\mathbf{x}),
\label{eq:sd_forward}
\\
\forall t,\ \underset{\theta}{\text{min}}\ \mathbb{E}||(\mathbf{\epsilon}-&\epsilon_{\theta}(\mathbf{x}^t;\mathbf{c},t))\odot\mathbf{m}||_2^2,
\label{eq:sd_train}
\end{align}
where $\alpha_t$, $\sigma_t$ are scalar functions that describe the magnitude of the data $\mathbf{x}^0$ and the noise $\epsilon$ at the denoising step $t$, $\theta$ parameterizes the denoiser $\epsilon_{\theta}$, and $\mathbf{c}$ is the map tensor guiding the denoising process.
As illustrated in \cref{fig:preliminary} (b), all agent tokens are iteratively generated from the standard Gaussian noise with a \textit{uniform} denoising step $t$ during sampling.
The full sequence inpainting enables goal-oriented generation, in which the model sets a keep mask $\mathbf{m}_{c}$ to ensure targets and past tokens remain fixed during sampling:
\begin{equation}
p(\mathbf{x}^{s}|\mathbf{x}^{t}) = \mathcal{N}(\mathbf{x}^{s}|\mu(\mathbf{x}^t,t),\Sigma(\mathbf{x}^t,t))\odot \bar{\mathbf{m}}_{c}+\mathbf{x}^{t} \odot \mathbf{m}_{c},
\label{eq:label}
\end{equation}
where $s$ is the next denoising step, $\mu$ and $\Sigma$ are determined by DiT $\epsilon_{\theta}$.
However, fixed-length denoising prevents intermediate state updates, making the model unable to react dynamically to environmental changes during generation.
%
%

\begin{figure*}[t]
\begin{center}
\centerline{\includegraphics[width=\linewidth]{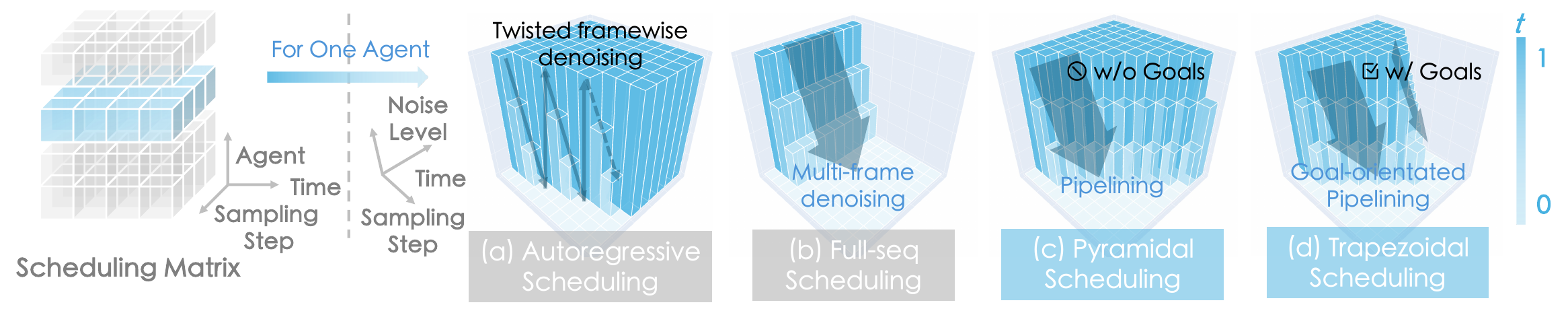}}
\caption{
\textbf{Diagram of the scheduling strategy.}
%
An agent's noise varies between zero and one across timesteps, determining the balance between stochasticity and goal-driven guidance at each sampling step.
%
%
\textbf{(a) }
is hindered by excessive steps per frame.
\textbf{(b) }
reduces costs and follows guidance but can't react to abrupt changes.
\textbf{(c) }
distributes cost by progressively adding tokens to the active chunk only at the start of each step, ensuring smoother transitions and better reactivity.
\textbf{(d) }
enhances future guidance and reduces cost by completing the path from both ends when the goal is fixed. The last two are our options.
}
\vspace{-30pt}
\label{fig:method}
\end{center}
\end{figure*}

\section{Nexus Framework}
\label{framework}

Nexus adaptively generates realistic driving scenarios by leveraging decoupled diffusion states for goal-oriented guidance and responsive scheduling.
%
The training stage begins with encoding the agent and the map into tokens with randomly added noise (\cref{fig:pipeline}).
For goal orientation, Nexus treats independent noise states as partial masks and uses a diffusion transformer to learn from low-noise guidance via sequence completion (\cref{sec:Next-chunk Diffusion}).
For reactivity, Nexus schedules tokens dynamically based on noise states during sampling, updating active elements continuously while removing generated ones to ensure timely scenario reaction (\cref{sec:scheduling}).
%
%
%
%
%
%
The training data for conditioned generation in safety-critical scenarios is presented in \cref{sec:Nexus-Data}.

\subsection{Noise-masking Training for Goal Orientation}
\label{sec:Next-chunk Diffusion}

Existing approaches~\cite{jiang2024scenediffuser} train diffusion models with \textit{uniform} noise, relying on hard-masked conditioning to inpaint missing scene components.
%
Yet, sampling requires continuous denoising of a fixed-length sequence to incorporate future guidance.
In response to updates, the model discards and regenerates upcoming parts, reducing flexibility.
%
%
%
%
%
Instead, we propose decoupled diffusion, where \textit{independent} noise levels act as soft masks, enabling Nexus to selectively follow low-noise goal tokens while flexibly reusing or skipping steps for efficient adaptation.

\noindent
\textbf{Noising as partial masking.}
Generative models are essentially various forms of mask modeling~\cite{chen2024diffusion}, which share 
%
the practice of occluding a subset of data and training a model to recover unmasked portions.
%
In particular, training full-sequence diffusion can be treated as \textit{noise-axis masking}, namely adding unified noise to the data $\mathbf{x}^0$ over a fixed-length sequence.
The sampling process gradually denoises $\mathbf{x}^t$ from Gaussian noise, with the mask being progressively removed.
While next-token prediction, which masks each token $\mathbf{x}_{\tau+1:}$ at $\tau$ and masks predictions from the past $\mathbf{x}_{:\tau}$, is a form of \textit{time-axis masking}.
Masked parts gradually reveal over time with no restrictions on length or composition.


%

We explore unifying the best of both by leveraging the noise states as partial masks across all dimensions to merge diffusion with next-token prediction.
According to~\cite{chen2024diffusion}, any collection of tokens can be viewed as an ordered set with unified indices along all axes without loss of generality.
Inspired by~\cite{chen2024diffusion}, we introduce tri-axial mask modeling, where independent noise levels align across agent indices, temporal timesteps, and denoising steps to unify diffusion with next-token prediction.
Specifically, we denote $\boldsymbol{x}_{a,\tau}^{k_{a,\tau}}$ as the token of $a$-th agent $\boldsymbol{x}_{a,\tau}$ within $\mathbf{x}^{k_{a,\tau}}$ at noise level $k_{a,\tau}$ under the forward diffusion process in \cref{eq:sd_forward}; $\boldsymbol{x}_{a,\tau}^0$ and $\boldsymbol{x}_{a,\tau}^T$ represent the unnoised token and the pure noise.
The noise level matrix $\mathbf{k} = [k_{a,\tau}]\in(0,1]^{A \times \mathcal{T}}$ of the sequence is assigned a random matrix, representing the degrees of Gaussian noise added to corresponding tokens.
%
%
The optimizing process of the scene generation model can be rewritten as:
\begin{equation}
    \forall ~\mathbf{k} \in (0,1]^{A\times \mathcal{T}},\ \underset{\theta}{\text{min}}\ \mathbb{E}||(\mathbf{\epsilon}-\epsilon_{\theta}(g(\mathbf{x}^0,\mathbf{k});\mathbf{c},\mathbf{k}))||_2^2,
\end{equation}
where $g$ represents the function that adds noise to $\mathbf{x}^0$ using matrix $\mathbf{k}$, where each token is masked to varying degrees.
The model is learned by completing the full sequence from soft-masked tokens, following information from low-noise tokens when generating other parts.
During sampling, setting history and goals to low noise and others to high noise ensures conditional guidance in scene generation.

\noindent
\textbf{Scene tokenizing and encoding.}
Nexus builds upon the diffusion transformer (DiT)~\cite{peebles2023scalable}, employing structured tokenization and encoding to provide a unified representation of driving scenarios.
In \cref{fig:pipeline} (a), the model first extracts vectorized map tensor $\mathbf{c}$ and agent tensor $\mathbf{x}$ from offline-collected driving logs (\cref{sec:preliminary}).
Each tensor is channel-normalized and encoded via MLP for unified processing of coordinates, size, speed, \textit{etc}.
To ensure stable learning, we initialize a set of learnable queries and use Perceiver IO~\cite{jaegle2021perceiver} to encode the map into fixed-length tokens.
After adding random noise to the agent tensor, a two-dimensional rotary positional embedding is applied to let the model have a sense of both physical time and denoising steps.

\noindent
\textbf{Modeling interactions with multiple attention.}
The design of Nexus's diffusion transformer focuses on the integration of agent-agent interactions with structured map-based reasoning, ensuring realistic coordination of trajectories and lane-following behaviors (\cref{fig:pipeline} (b)).
Firstly, a map cross-attention queries the map using the agent tensor, aiding in agent-map interactions like lane following and merging.
%
%
Then, the agent tensor is used to condition a set of temporal and spatial transformer blocks via AdaLN~\cite{peebles2023scalable}.
It captures trajectory continuity and spatial interactions like following and yielding.
Besides, the validity $\mathbf{m}$ is used as an attention mask within the transformer denoiser, and invalid and skipped tokens outside the chunk are excluded.
The final MLP decodes the agent tokens to compute the reconstruction loss against the ground truth.

\subsection{Noise-aware Scheduling for Reactivity}
\label{sec:scheduling}
After training, Nexus defines the chunk as a localized subset of the scenario, where varying noise states guide the model to prioritize low-noise cues during denoising.
To optimize reactivity, we introduce a noise-aware scheduling strategy that arranges the denoising sequence of scene components for real-time adaption to environmental updates.

\noindent
\textbf{Scene generation from next-chunk prediction.}
Nexus structures each chunk with historical context, future frames to be denoised, and optional goal tokens while adjusting noise levels dynamically at each denoising step.
As shown in \cref{fig:pipeline} (c), the chunk includes frames at time $\tau$, $\tau+1$, and goals, with the darker shade of blue indicating higher noise.
After denoising at $t$, all tokens' noise levels drop. 
The denoised token at $\tau$ is popped out, and a high-noise frame at $\tau+2$ is pushed into the chunk for the next denoising step $s$. 
Any environmental changes to the agent tensor can replace the agent state directly and reduce the noise.
Thus, the chunk slides temporally as tokens are updated.

We define the scheduling process as a three-dimensional matrix $\mathcal{K}\in[\mathbf{k}]^{M}$ (\cref{fig:method} left), where each entry encodes the noise level $t$ for an agent $a$ at physical timestep $\tau$ during sampling step $m$.
The sequence is initialized with white noise and the scheduling matrix $\mathcal{K}$ as 1.
The noise level of historical frames and the optional goal is set to 0.
Nexus selects the noise level $\mathcal{K}_{m}$ along the sampling step indice $m$. 
\cref{fig:method} (right) shows a fixed agent's noise level changes over sampling steps (height and color), with arrows indicating denoising.
Tokens enter the chunk when their noise level changes and exit when it reaches zero, repeating until the entire sequence is generated.

\noindent
\textbf{Scheduling for pipelined generation.}
Tokens with different noise levels are strategically scheduled with matrix $\mathcal{K}$, enabling the model to follow low-noise tokens for denoising without retraining.
However, naive autoregressive scheduling is slow due to twisted frame-by-frame denoising, while full-sequence generation, though faster, lacks reactivity to changes.
Therefore, we use a pipelined strategy to denoise multiple frames simultaneously, distributing the cost within the chunk.
In pyramidal scheduling, the chunk length scales with the number of denoising steps. Each iteration introduces new frame tokens while removing fully denoised ones, allowing continuous scene refinement and output once the chunk is saturated.
Trapezoidal scheduling enables bidirectional token updates, where tokens enter and exit from both ends of the chunk, reinforcing goal-conditioned trajectory synthesis.
%
The goal's guidance can propagate to other agents through spatial-temporal attention within the Nexus.
%

%
%
%

%

\noindent
\textbf{Behavior alignment via classifier guidance.}
%
%
To align generated scenes with realistic driving behavior, we incorporate classifier guidance inspired by dynamic thresholding~\cite{saharia2022photorealistic}, adjusting noise levels at each step to refine agent interactions.
%
We refine \cref{eq:label} by applying $f(\mathbf{x}^t, t)$ as a corrective function, adjusting agent trajectories iteratively to enforce behavior constraints at each denoising step.  
Practically, we separate overlapping agents along their centerline's opposite direction to avoid collisions, smooth trajectories, and pull agents toward the nearest lanes for on-road driving.
The formula detail is provided in Appendix {\color{red}C.3}.


\begin{figure}[t]
\begin{center}
\centerline{\includegraphics[width=\linewidth]{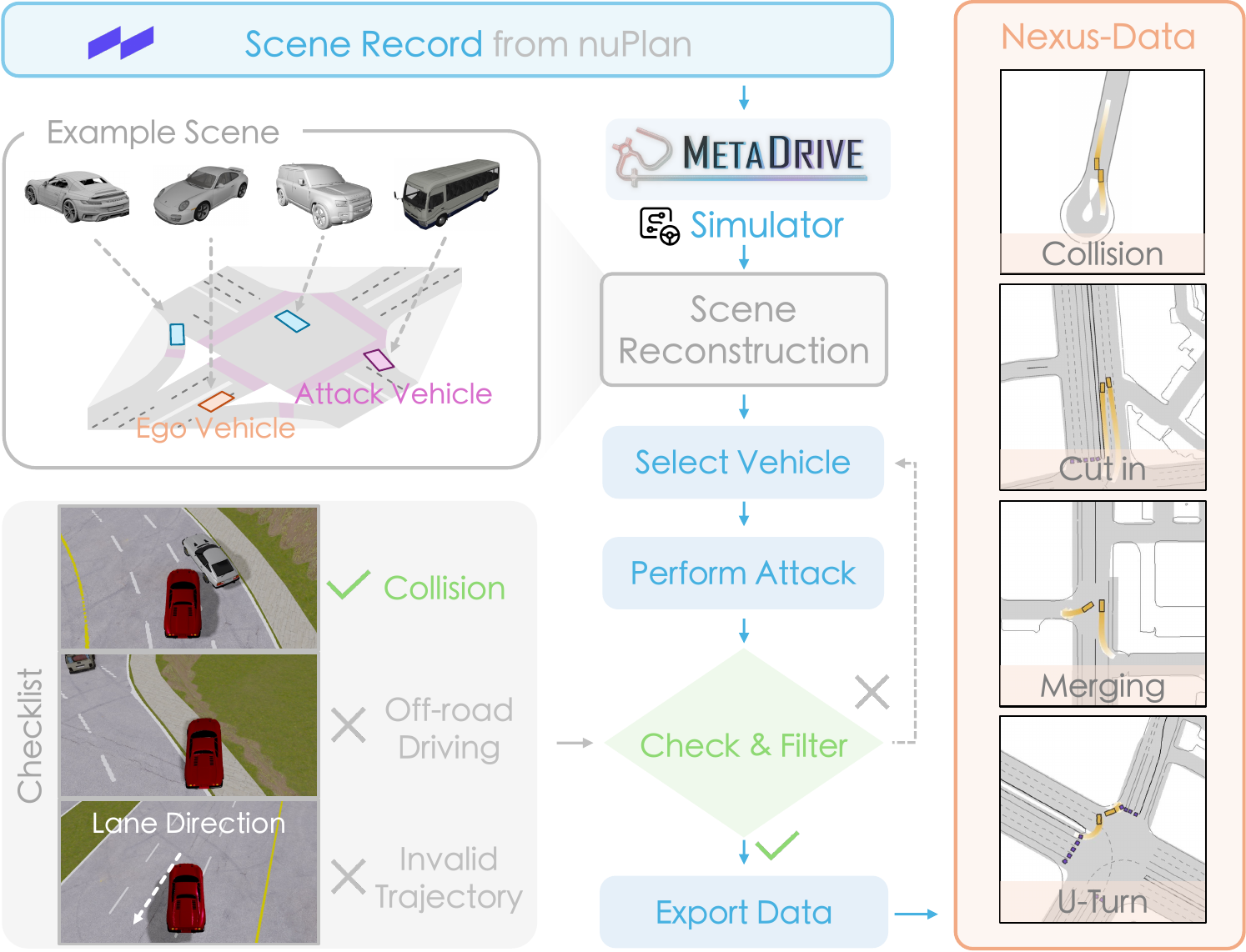}}
\caption{
\textbf{Nexus-Data construction.}
Nexus-Data employs scene records from the nuPlan dataset to reconstruct maps and agents in a simulator to ensure scene realism.
It selects a neighbor vehicle to generate attack trajectories by adversarial learning~\cite{zhang2023cat} and filters out unrealistic cases.
}
\vspace{-30pt}
\label{fig:nexus-data}
\end{center}
\end{figure}

\begin{figure*}[t]
\centering
\includegraphics[width=\linewidth]{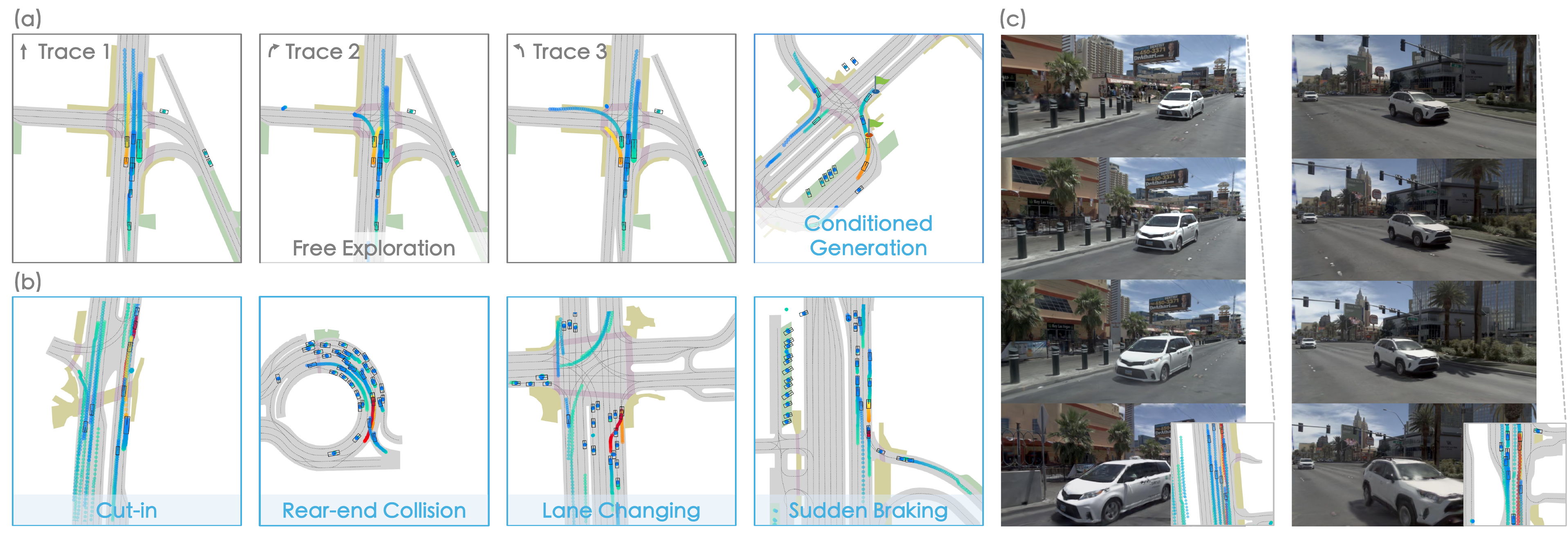}
\caption{
\textbf{Visualization of Nexus.}
\textbf{(a)} Free exploration generates diverse future scenarios from initialized history, while conditioned generation synthesizes scenes based on predefined goal points.
\textbf{(b)} Setting the attacker's goal as the ego's waypoint enables adversarial scenarios.
\textbf{(c)} Neural radiance fields provide Nexus-generated scenes with realistic, risky driving visuals.
Color transitions represent motion. Ego: yellow to orange. Attacker: red to magenta. Others: green to blue.
}
\vspace{-10pt}
\label{fig:main_vis}
\end{figure*}

\subsection{Nexus-Data for Generalization in Risky Scenes}
\label{sec:Nexus-Data}

Existing datasets predominantly feature safe driving behaviors, limiting exposure to rare corner cases and leading to trajectory discontinuities.
To address this, we construct \textbf{Nexus-Data}, using MetaDrive~\cite{li2022metadrive}, to enhance generalization in safety-critical scenarios by changing the motions of the logged agents with adversarial traffic generation~\cite{zhang2023cat}.
%

\noindent \textbf{Scene layout construction.}
We utilize ScenarioNet~\cite{li2023scenarionet} to transform scenes into a unified description format suitable for simulators, known as \textit{scene records}, logging the agent and map information.
As illustrated by the example scene in \cref{fig:nexus-data}, loading \textit{scene records}, MetaDrive~\cite{li2022metadrive} can reconstruct lanes, roadblocks, and intersections and place corresponding 3D models based on the recorded positions and orientations.
%
%
By doing so, the digital twin scenario can be faithfully reconstructed in the simulator.

%
%
%

\noindent \textbf{Creation of safety-critical data.} 
As collisions naturally lead to hazardous situations, we use CAT~\cite{zhang2023cat} to generate risky scenes initialized from real-world layouts.
Specifically, a candidate pool with distance $d$ of the ego vehicle is defined, and we randomly select one as the attack vehicle.
Using adversarial learning~\cite{zhang2023cat}, we generate high-risk interactions by selecting the most collision-prone trajectory for the attack vehicle, forcing the ego vehicle to execute avoidance maneuvers under realistic constraints.
After $K$ rounds, the attack trajectory is generated.
Despite full-stack automation, our statistical study shows that only 36.9\% of scenarios result in valid collisions, let alone reasonable ones.
Thus, we introduce a checklist to filter out non-collision cases, off-road driving, and invalid trajectories (\textit{e.g.}, in-place U-turns, and lateral shifts), using assert mesh detection and rules.
%
%
If no valid attacker is found, we continue iterating until the pool is depleted.
Eventually, we collect 540 hours of high-quality driving scenarios, covering risky behaviors like sudden braking, crossroad meeting, merging, and sharp turning.
%

\section{Experiments}

\noindent
\textbf{Setup and protocols.}
Nexus is trained on nuPlan~\cite{caesar2021nuplan}, Waymo~\cite{sun2020scalability}, and our self-collected Nexus-Data, ensuring exposure to both standard and safety-critical driving scenarios.
The performance is evaluated based on controllability, interactivity, and kinematics.
We assess trajectory accuracy using average displacement error (ADE),
measuring the deviations from ground truth trajectories.
The offroad rate measures the proportion of agents that deviate from the centerline beyond a threshold.
Collision rate quantifies the safety of generated trajectories, assessing how well Nexus models interactions between agents while avoiding crashes.
Metrics related to velocity and angular change describe the trajectory instability of agents' movement.
%

\begin{table}[]
\centering
\footnotesize
\caption{\textbf{Generation controllability, interactivity, and kinematics compared to nuPlan experts.}
The tasks predict 8-second futures from 2-second history, with or without a goal. \texttt{ADE}: displacement error, $R_{\text{road}}$ and $R_{\text{col}}$: off-road and collision rate (\%), $M_{\text{k}}$: instability.
\textit{Full}: model with Nexus-Data and classifier guidance.
\colorbox{baselinecolor}{gray}: main metric. \textbf{bold}: best results.
}
\tablestyle{2.0pt}{1.05}
\begin{tabular}{l|cccc|cc|c}
\toprule
\multirow{2}{*}{Method} & \multicolumn{4}{c|}{Conditioned Generation}      & \multicolumn{2}{c|}{Free Exploration} & Time \\  
& \baseline{ADE $^\downarrow$} & $R_{\text{road}}$ $^\downarrow$ & $R_{\text{col}}$ $^\downarrow$ & $M_{\text{k}}$ $^\downarrow$ & $R_{\text{col}}$ $^\downarrow$      & $M_{\text{k}}$ $^\downarrow$ &  (Sec)\\ \midrule
\textit{\color{gray}{IDM~\cite{treiber2000congested}}}      &   \baseline{\color{gray}{10.52}}       &   \color{gray}{9.85}      &  \color{gray}{10.17}           &   \color{gray}{6.30}    &   \color{gray}{12.10}    &     \color{gray}{6.02}   & \color{gray}{2.16}   \\ \midrule
D. Policy~\cite{chi2023diffusion}      &    \baseline{7.80}     &  13.9       &     14.92        & 12.71       &    16.88   & 9.30     & 6.59     \\
SceneD.~\cite{qian2023nuscenes}      &    \baseline{5.99}     &    8.53     &     11.78        &   9.64    &    13.59   &     6.16   &  5.34  \\
GUMP~\cite{hu2025solving}      &   \baseline{1.93}       &   7.73      &      7.85       &     16.18  &   10.23    &      14.30   & 5.59  \\
 \midrule
Nexus         &  \baseline{1.28}   &  6.89       &   1.62          &   4.63    &   2.61    &  3.23    &   \textbf{2.79}   \\
\textbf{Nexus-\textit{Full}}          &  \baseline{\textbf{1.12}}   &   \textbf{6.25}      &      \textbf{1.56}     &    \textbf{3.17}   &    \textbf{2.10}   &     \textbf{2.14}  & 2.93    \\ \bottomrule
\end{tabular}
\vspace{-10pt}
\label{tab:nuplan_eval}
\end{table}

\subsection{Comparison to State-of-the-arts}

%
We compare Nexus with the recently available and reproduced data generation approaches trained on the nuPlan dataset.
Conditioned generation and free exploration tasks are used for evaluation. 
The former probabilistically provides a goal point, while the latter generates an 8-second future freely, with both using the first two seconds as context.
\cref{tab:nuplan_eval} shows that Nexus surpasses all previous methods in controllability (ADE), interactivity ($R_{\text{road}}$ and $R_{\text{col}}$), and kinematics ($M_{\text{k}}$).
Specifically, Nexus significantly improves ADE by \textbf{-4.71} compared to SceneDiffuser~\cite{jiang2024scenediffuser} while reducing generation time by \textbf{-2.55} seconds.
It also excels in collision rate (\textbf{1.56\%}) and trajectory instability, showcasing Nexus's multi-agent interaction modeling.
\cref{fig:main_vis} (a) shows Nexus generating diverse, realistic futures (trace 1-3) from the same initial conditions. 
Besides, it highlights Nexus's goal adherence and controllability in the conditioned generation for driving simulation.

To boost controllability and corner case generation, we introduce Nexus-\textit{Full}, which incorporates extra Nexus-Data and classifier guidance.
This improves scene controllability, reducing ADE by \textbf{-0.18} while maintaining interactive realism with minimal time increase from guidance.
Nexus-\textit{Full} exhibits strong generalization
ability across scenarios (\cref{fig:main_vis} (b)).
It covers safety-critical driving behaviors, including cut-in, collision, lane changing, \textit{etc}.
 
Beyond nuPlan, we also evaluate the unconditional generation of Nexus on the Waymo Open Sim Agent \textit{val} set, scoring \textbf{61.9} on the composite metric without any post-processing, outperforming the state-of-the-art competitor SceneDiffuser~\cite{jiang2024scenediffuser} (55.8).


\begin{table}[]
\centering
\footnotesize
\caption{
\textbf{Comparisons on scheduling strategies.}
In addition to task performance, we report the sampling steps, reaction time to changes, and overall time to generate an 8-second scene.
\texttt{A.R.}: autoregressive.
$\dagger$: updating histories by receiving interactions.
}
\tablestyle{2.0pt}{1.05}
\begin{tabular}{l|cccc|ccc}
\toprule
\multirow{2}{*}{Scheduling} & \multicolumn{4}{c|}{Conditioned Generation} & \multirow{2}{*}{Steps} & React      & Overall \\  
& \baseline{ADE $^\downarrow$} & $R_{\text{road}}$ $^\downarrow$ & $R_{\text{col}}$ $^\downarrow$ & $M_{\text{k}}$ $^\downarrow$ &       & Time & Time  \\ \midrule
A.R.      &   \baseline{1.48}       &   9.95      &  1.98           &   \textbf{4.58}    &   512    &     4.96   & 79.36   \\ 
Full-sequence      &    \baseline{1.28}     &  9.63       &     1.62        & 4.63       &    \textbf{32}   & 4.96     & \textbf{4.96}     \\ \midrule
Pyramidal      &   \baseline{1.53}       &   9.85      &      1.80       &     4.74  &   48    &   \textbf{0.16}   &   7.68   \\
Trapezoidal      &    \baseline{1.39}     &    9.70     &     1.92        &   4.63    &    40   &     \textbf{0.16}   &  6.20  \\
\textbf{Trapezoidal $^\dagger$}         &  \baseline{\textbf{1.17}}   &  \textbf{9.54}       &   \textbf{1.71}          &   4.89    &   40    &  \textbf{0.16}    &   6.20   \\ \bottomrule
\end{tabular}
\label{tab:scheduling_ablation}
\end{table}

\begin{table}[]
\centering
\footnotesize
\caption{
\textbf{Ablation on designs in Nexus.}
\texttt{P.E.}: positional embedding with physical and denoising time.
All proposed designs contribute to the final performance.
}
\tablestyle{2.0pt}{1.05}
\begin{tabular}{l|cccc|cc}
\toprule
\multirow{2}{*}{Method} & \multicolumn{4}{c|}{Conditioned Generation}      & \multicolumn{2}{c}{Free Exploration}  \\  
& \baseline{ADE $^\downarrow$} & $R_{\text{road}}$ $^\downarrow$ & $R_{\text{col}}$ $^\downarrow$ & $M_{\text{k}}$ $^\downarrow$ & $R_{\text{col}}$ $^\downarrow$      & $M_{\text{k}}$ $^\downarrow$   \\ \midrule
Baseline      &   \baseline{7.53}       &   9.74      &  13.52           &   11.47    &   15.79    &     6.02     \\ \midrule
+ Noise Masking      &    \baseline{3.42}     &  9.16       &     3.01        & 8.19       &    3.48   & 5.23          \\
+ P.E.      &   \baseline{1.44}       &   8.17      &      2.52       &     6.20  &   3.17    &    3.42      \\
+ Nexus-Data      &    \baseline{1.32}     &   7.53     &     1.92        &   4.87    &    2.76   &     3.35    \\
+ Classifier Guidance         &  \baseline{\textbf{1.25}}   &  \textbf{6.73}       &   \textbf{1.47}          &   \textbf{4.02}    &   \textbf{1.38}    &  \textbf{3.28}       \\ \bottomrule
\end{tabular}
\vspace{-10pt}
\label{tab:ablation}
\end{table}

\subsection{Ablation Study}

\noindent
\textbf{Comparison on scheduling strategies.}
The ablation is conducted by training each variant of our model on nuPlan with 30K steps.
\cref{tab:scheduling_ablation} compares the impact of different scheduling strategies of Nexus on performance and speed.
Traditional scheduling strategies introduce significant latency, requiring \textbf{4.96} seconds to respond to environmental changes, making them impractical for real-time scene adaptation.
Notably, the pyramidal and trapezoidal scheduling strategies respond to changes during each denoising step, reducing response time by \textbf{-4.80} seconds without performance loss. 
When Nexus updates the historical state in real-time using feedback from agents, ADE improves by \textbf{+0.11} compared to full-sequence scheduling. 
Further, by applying step skipping~\cite{song2023consistency}, the sampling steps can be reduced to 18, generating 8-second future scenes in just \textbf{2.79} seconds.

\begin{figure}[t]
\centering
\includegraphics[width=1\linewidth]{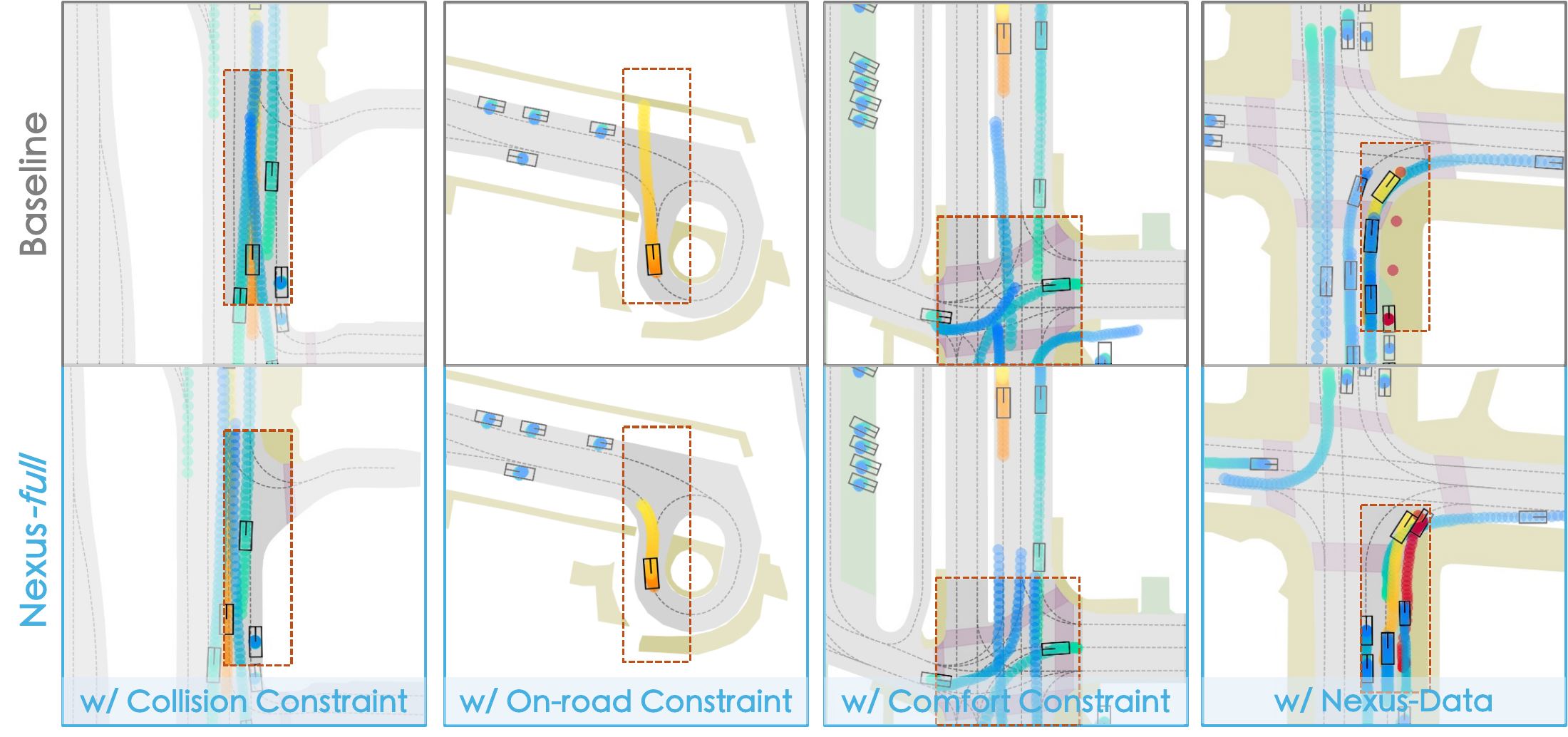}
\caption{
\textbf{Ablation on classifier guidance and Nexus-Data.}
The designs improve collision avoidance, on-road driving, stability, and corner case generation. Yellow is the ego agent, blue is the others, and red is the attacker.
}
\vspace{-10pt}
\label{fig:ablation-vis}
\end{figure}

\noindent
\textbf{Ablation on each component.}
To validate each component's effectiveness, we gradually introduce our proposed components and conditions, starting with a Diffusion Policy baseline~\cite{chi2023diffusion}.
As \cref{tab:ablation} shows, noise-masking training with independent noise states reduces ADE by \textbf{-4.11}, enabling the model to follow low-noise cues via sequence reconstruction.
Likewise, encoding physical and denoising time boosts performance by \textbf{-1.98}, aligning with our independent noise design.
Integrating Nexus-Data improves ADE by \textbf{-0.12}, enhancing controllability in diverse scenarios. 
%
Lastly, adding classifier guidance from human behavior constraints boosts collision (\textbf{-0.45\%}) and kinematic metrics (\textbf{-0.85}).
\cref{fig:ablation-vis} shows how Nexus-Data enhances corner case generation. 
Besides, adding constraints improves safe distancing, on-road driving, and trajectory stability.

\subsection{Discussion on Application}

\noindent
\textbf{World generator for closed-loop driving.}
Nexus enables closed-loop scene generation, acting as an interactive environment for autonomous agents.
The agent uses generated scenes for planning, while Nexus updates them in real-time based on the agent's actions.
To assess the realism of the scene generator, we set up an evaluation on the nuPlan closed-loop Val14 set (\cref{tab:world_model}).
The generator predicts the next scenario using history and agent actions. 
A baseline agent, Diffusion Planner~\cite{zheng2025diffusion}, predicts future waypoints based on the generated scenes, with more realistic scenes yielding higher metrics in the nuPlan closed-loop evaluation.
Nexus surpasses all baselines in reactive closed-loop evaluation by \textbf{+15.8}, highlighting its ability to generate interactive, realistic driving environments for closed-loop planning and policy learning.


\noindent
\textbf{Data augmentation via synthetic data.}
Nexus can generate diverse future scenarios from fixed history, making it a possible data engine for augmenting planning model training.
%
%
In a preliminary experiment, we sample 3 hours of nuPlan logs, generate synthetic data with Nexus, and train a lightweight planner~\cite{chi2023diffusion} on real and augmented data. 
\cref{tab:data_engine} indicates that blending generated data with real-world data improves reactive closed-loop score to \textbf{57.86}, a \textbf{+20\%} improvement over real-data-only models. 
This demonstrates that high-quality synthetic data can enhance planner robustness and generalization.
Models trained on small real datasets tend to slow down, raising collision scores while worsening other metrics.
We also find that limited synthetic data (3$\times$) degrades performance, likely due to scene noise affecting planner learning. 
A 30$\times$ increase brings significant gains, but further expansion leads to model saturation.
This shows that sufficient data scaling benefits the driving model, further validating Nexus as a reliable synthetic data generator for driving model training.


\begin{table}[t]
\centering
\footnotesize
\caption{
\textbf{Evaluation of a generation model as a world generator.}
The scene generator serves as the interactive world model response to the baseline planner's actions, with nuPlan closed-loop metrics reflecting its realism.
\texttt{Oracle}: ground truth environment of nuPlan evaluation.
$S_{\text{col}}$: collision score, $S_{\text{p}}$: progress score.
}
\tablestyle{2.0pt}{1.05}
\begin{tabular}{l|ccc|ccc}
\toprule
\multirow{2}{*}{Method} & \multicolumn{3}{c|}{Reactive Eval.}      & \multicolumn{3}{c}{Non-reactive Eval.}  \\  
& \baseline{Score $^\uparrow$} & $S_{\text{col}}$ $^\uparrow$ & $S_{\text{p}}$ $^\uparrow$ & Score $^\uparrow$ & $S_{\text{col}}$ $^\uparrow$      & $S_{\text{p}}$ $^\uparrow$   \\ \midrule
\color{gray}{Oracle}      &   \baseline{\color{gray}{82.8}}       &   \color{gray}{89.5}      &  \color{gray}{97.0}           &   \color{gray}{89.2}    &   \color{gray}{91.6}    &     \color{gray}{100.0}     \\ \midrule
Diffusion Policy~\cite{chi2023diffusion}      &    \baseline{61.6 \tiny\textcolor{DarkBlue}{(-21.2)}}     &  81.9       &     90.2        & 47.2 \tiny\textcolor{DarkBlue}{(-42.0)}       &    67.0   & 89.8          \\
SceneDiffuser~\cite{jiang2024scenediffuser}      &   \baseline{57.2 \tiny\textcolor{DarkBlue}{(-25.6)}}       &   74.7      &      91.6       &     50.1 \tiny\textcolor{DarkBlue}{(-39.1)}  &    66.3   &   89.5       \\
\textbf{Nexus}      &    \baseline{\textbf{73.0 \tiny\textcolor{DarkBlue}{(-9.8)}}}     &    \textbf{84.9}     &     \textbf{95.0}        &   \textbf{68.1 \tiny\textcolor{DarkBlue}{(-21.1)}}    &    \textbf{77.7}   &     \textbf{96.6}    \\ \bottomrule
\end{tabular}
\vspace{-10pt}
\label{tab:world_model}
\end{table}

\noindent
\textbf{Attempts to visual world models.}
Nexus is not designed for visual synthesis.
Yet, high-quality traffic layout generation opens new possibilities for controllable visual scene rendering, which is essential for realistic interactive driving simulations and closed-loop testing.
We have an initial attempt to render nuPlan driving scenes using neural radiance fields.
We modify an open-source autonomous driving neural reconstruction method~\cite{tonderski2024neurad} to support Nexus-generated scenes, allowing control over agent positions and behaviors through novel layouts.
This shows a promising result of action-conditioned video generation for safety-critical scenarios in \cref{fig:main_vis} (c).
This application is impossible with existing world models, which are only trained and conditioned on a given static real-world dataset that lacks records of dangerous driving behaviors.
This brings new opportunities for closed-loop data generation capabilities.


\vspace{-5pt}
\section{Related Work}
\label{sec:related work}
\vspace{-5pt}

\noindent
\textbf{Diffusion models for conditioned generation.}
Diffusion models have advanced policy generation~\cite{chi2023diffusion,janner2022planning,liu2025skill,mishra2023generative} and scene synthesis~\cite{huang2023diffusion,li2024sat2scene}.
Some works explore their use in ego-motion planning~\cite{huang2024gen,jiang2023motiondiffuser,liao2024diffusiondrive,sun2023large}, while Diffusion-ES~\cite{yang2024diffusion} integrates evolutionary search for optimizing non-differentiable trajectories. 
Diffusion Planner~\cite{zheng2025diffusion} jointly predicts ego and surrounding vehicle trajectories, merging motion prediction with closed-loop planning. 
Other efforts focus on full-scene generation as a world generator~\cite{chitta2025sledge} or controllable traffic simulation via user-defined trajectories~\cite{zhong2023guided}. 
SceneDiffuser~\cite{jiang2024scenediffuser} refines diffusion denoising for efficient simulation with hard constraints and LLM-driven scene generation. 
However, these approaches rely on static dataset layouts, limiting their reactivity to dynamically evolving traffic conditions and safety-critical scenario synthesis. Nexus overcomes this by integrating decoupled diffusion with noise-aware scheduling for real-time reaction.
Despite amortized optimizations, existing diffusion models struggle with goal-oriented planning due to uniform noise treatment, which hinders precise control over agent trajectories. Additionally, their fixed-sequence denoising process limits timely reactivity to environmental changes.

\noindent
\textbf{Transformers for reactive generation.}
Closed-loop agent state prediction requires reaction mechanisms for simulation dynamics and realism~\cite{feng2023trafficgen,guo2023scenedm,qian20232nd,tan2023lctgen}. 
Decisions are influenced by the historical and current system state, including other agents' actions. 
MotionLM~\cite{seff2023motionlm} and TrajEnglish~\cite{philion2024trajeglish} use autoregressive GPT-like models, while GUMP~\cite{hu2025solving} improves response speed with key-value pair tokenization and state-space quantization, enabling flexible handling of agent disappearance and emergence.
Unlike autoregressive models, which rely on discrete updates, Nexus enables real-time scene adaptation by integrating noise-aware scheduling with goal-conditioned diffusion.

\noindent
\textbf{Conventional traffic simulation.}
Rule-based traffic simulation methods have demonstrated success in controlled settings~\cite{dauner2023parting,fan2018baidu,sun2022shift,treiber2000congested}
, but their rigid structure prevents them from adapting to novel interactions or emergent behaviors.
While imitation learning methods~\cite{chen2023end,chitta2022transfuser,hu2023_uniad} excel in behavior cloning, they lack adaptability to unseen scenarios, as their reliance on predefined rules leads to brittle, unsafe outputs in edge cases~\cite{huang2023gameformer,ljungbergh2024neuroncap,vitelli2022safetynet}. 
In contrast, Nexus learns flexible, data-driven representations that capture diverse driving behaviors, generalizing beyond predefined rule sets and enabling controllable scene generation.

\begin{table}[t]
\centering
\footnotesize
\caption{
\textbf{Comparison involving data augmentation using synthetic data.}
Nexus serves as a data engine, expanding sampled scenes to train the planner~\cite{chi2023diffusion} at varying scales. The nuPlan closed-loop evaluation demonstrates the performance gains from data augmentation. \texttt{Synth.}: synthetic.
}
\tablestyle{2.0pt}{1.05}
\begin{tabular}{l|ccc|ccc}
\toprule
Training & \multicolumn{3}{c|}{Reactive Eval.}      & \multicolumn{3}{c}{Non-reactive Eval.}  \\  
Data & \baseline{Score $^\uparrow$} & $S_{\text{col}}$ $^\uparrow$ & $S_{\text{p}}$ $^\uparrow$ & Score $^\uparrow$ & $S_{\text{col}}$ $^\uparrow$      & $S_{\text{p}}$ $^\uparrow$  \\ \midrule
Real Data      &   \baseline{48.11}       &   \textbf{80.81}      &  83.41           &   46.39    &   72.54    &     88.80     \\ \midrule
w/ 3$\times$ Synth. Data      &    \baseline{46.61}     &  75.78       &     86.25        & 43.69       &    66.93   & 89.15          \\
w/ 30$\times$ Synth. Data      &   \baseline{56.46}       &   78.92      &      \textbf{92.55}       &     53.39  &   69.86    &    \textbf{94.55}      \\
w/ 60$\times$ Synth. Data      &    \baseline{\textbf{57.86}}     &    79.50     &     91.96       &   \textbf{56.42}    &    \textbf{76.49}   &     93.32    \\ \bottomrule
\end{tabular}
\vspace{-10pt}
\label{tab:data_engine}
\end{table}

\vspace{-5pt}
\section{Conclusion}
\vspace{-5pt}

We introduce Nexus, a decoupled diffusion model that generates diverse driving scenarios using independent noise states for real-time reaction and goal-oriented control.
This enables dynamic evolution while maintaining precise trajectory conditioning.
%
We curate Nexus-Data, a large dataset of safety-critical scenarios,
helping Nexus better model edge cases and high-risk interactions.
%
By integrating noise-masking training and noise-aware scheduling, Nexus achieves goal-driven scene synthesis with improved fidelity and diversity, improving autonomous driving simulations.

\noindent
\textbf{Limitations and future work.}
Nexus currently focuses on structured layout generation but does not jointly synthesize videos, limiting its applicability to closed-loop training of end-to-end driving. 
Extending Nexus to video-based generation and continuous online learning is left for future work.
%

\section*{Acknowledgements}

We extend our gratitude to Shenyuan Gao, Li Chen, Chonghao Sima, Carl Lindstr\"{o}m, Adam Tonderski, and the rest of the members from OpenDriveLab and Zenseact for their profound discussions and supportive help in writing.


{\small
\bibliographystyle{ieee_fullname}
\bibliography{short,paper}
}

\onecolumn
\maketitle

\appendix

\noindent\textbf{\Large{Appendix}}

\startcontents
{
    \hypersetup{linkcolor=black}
    \setlength{\parskip}{1.5ex}
    \printcontents{}{1}{}
}
\newpage

\section{Discussions}

To better understand our work, we supplement with the following question-answering.

\textbf{Q1.}
\textit{What makes Nexus stand out compared to driving simulators?}

Current simulators~\cite{dosovitskiy2017carla,li2022metadrive} rely on hand-crafted rules, thus struggling with complex, out-of-scope scenarios. 
Generating corner cases requires manually positioning attack vehicles and adjusting traffic responses, making large-scale closed-loop adversarial scene generation impractical. 
While adversarial attacks~\cite{zhang2023cat} can create scenarios, log-replayed environmental agents lack realism and fail to ensure attack validity.
In contrast, Nexus proposes a scalable, user-friendly approach for realistic and controllable hazard scenario generation. 
It leverages diffusion models to capture vehicle interactions, ensuring realism. 
Our method requires only goal points for the ego and attack vehicles, easily defined as lane center points, enabling efficient large-scale scenario expansion. 
Details of scene generation are in \cref{sec:novel scenario}.

\vspace{5pt}

\textbf{Q2.}
\textit{What is the definition of safety-critical scenarios and how to ensure they are realistic and feasible?}

\textbf{Defination.}
A safety-critical scenario is a situation where one or more vehicles collide with the ego vehicle, which is rare to collect in real-world datasets like nuPlan. 
We utilize CAT~\cite{zhang2023cat} to generate risky behaviors from logged scenarios to ensure the reality and feasibility of training data, which uses a data-driven motion prediction model that predicts several modes of possible trajectories of each traffic vehicle. 
Please refer back to \cite{zhang2023cat} for a detailed description of safety-critical scenarios.

\textbf{Rationality of goal conditioning.}
Nexus emphasizes goal-controlled scenario generation, as it enables the convenient and scalable creation of \textit{collision-prone} corner cases. 
Given the trajectory of the target agent, an attack can be easily executed by setting the attacker's goal to a future waypoint of the target agent.
Nexus's design incorporates scenario interactions to enhance the realism of collisions.

\textbf{Evaluation.}
Evaluating the quality of generated corner cases scientifically is a well-recognized challenge in academia.
Our preliminary attempt combines quantitative and qualitative assessments.
We use goal-driven kinematic metrics to measure trajectory authenticity, where Nexus excels, and ensure generated scenarios meet industry-standard corner cases~\cite{favaro2023building,thorn2018framework} like cut-ins, sudden braking, and collisions (see \cref{fig:supp-attack-vis} for more visualizations).

\vspace{5pt}

\textbf{Q3. Broader impact.}
\textit{What are potential applications and future directions with the provided Nexus-data and the Nexus model, for both academia and industry?}

\textbf{Datasets.}
Nexus-Data collects massive data from simulators, significantly enhancing the layout diversity of driving scenarios. 
This dataset provides the community with high-quality resources for studying complex agent interactions, multi-agent coordination, and safety-critical decision-making in autonomous driving.

\textbf{Models.}
Beyond data augmentation, we believe our model can also drive broader applications within the community. 
This work showcases Nexus's potential as both a closed-loop world generator and a data engine. 
It could be adapted for downstream tasks, such as closed-loop training of autonomous driving agents~\cite{peng2022reward}.
Our model presents a promising generative world model, providing an alternative to traditional rule-based simulators.
Please note that our model will be publicly released to benefit the community and can be further fine-tuned flexibly according to custom data within the industry.

\textbf{Negative societal impacts.}
The potential downside of Nexus could be its unintended use in generating counterfeit driving scenarios due to the hallucination issues that may arise with diffusion models.
We plan to introduce rule-based validation mechanisms, such as collision consistency checks, kinematic feasibility constraints, and behavioral plausibility tests, to filter out unrealistic generated scenarios.
Besides, we plan to regulate the effective use of the model and mitigate possible societal impacts through gated model releases and monitoring mechanisms for misuse.


\vspace{5pt}

\textbf{Q4. Limitations.}
\textit{What are the issues with current designs and corresponding preliminary solutions?}

Visual synthesis is necessary for current end-to-end models in autonomous driving.
Yet, datasets with visual data~\cite{qian2023nuscenes} are still much less abundant compared to those containing only driving logs~\cite{caesar2021nuplan}. 
Nexus currently lacks visual generation, limiting its use in applications requiring realistic sensor data, such as perception model training and end-to-end learning for autonomous vehicles.

However, as a work exploring how to incorporate world generators with generative models, the primary focus of this work is the decoupled diffusion for adaptive scene generation. 
Future work may integrate Nexus with neural radiance fields (NeRFs)\cite{mildenhall2021nerf} for high-fidelity 3D scene synthesis or video diffusion models\cite{blattmann2023stable,yang2024cogvideox} for temporally consistent video generation, enabling full visual simulation of dynamic driving scenarios.
%
This would allow Nexus to generate scenarios with both rich agent behaviors and realistic visual information, improving the training of end-to-end models as a world model.

\section{Nexus-Data}
\label{sec:supp-nexus-data}


\begin{table}[]
\centering
\caption{
\textbf{Behavior distribution statistics.}
Proportion (\%) of agent behaviors in the dataset, excluding keeping forward.  
Our collected data provides a more balanced distribution for lane changes.
}
\tablestyle{2.0pt}{1.05}
\begin{tabular}{l|c|ccccccc}
\toprule
\multirow{2}{*}{Dataset} & Time    & Inter. & Left & Right  & L. Lane & R. Lane &  \multirow{2}{*}{U-Turn} & \multirow{2}{*}{Stop} \\  
& (Hrs)  & Passing & Turn & Turn & Change & Change & &  \\ \midrule
nuScenes~\cite{caesar2019nuscenes}      &    5.5     &  13.1      &     18.0        & 10.2       &   5.0   & 2.5 & 0.0  & 4.1         \\
nuPlan~\cite{caesar2021nuplan}      &   1.2K       &   13.8      &      1.5       &     1.6  &   14.4    &    14.6 & 0.9  &   46.8      \\
\textbf{Nexus-Data}      &    540     &    35.3     &     1.7        &   2.5   &    22.2   &    23.3  & 1.2 & 10.0    \\ \bottomrule
\end{tabular}
\label{tab:data_eval}
\end{table}

\begin{figure}
\centering
\includegraphics[width=\linewidth]{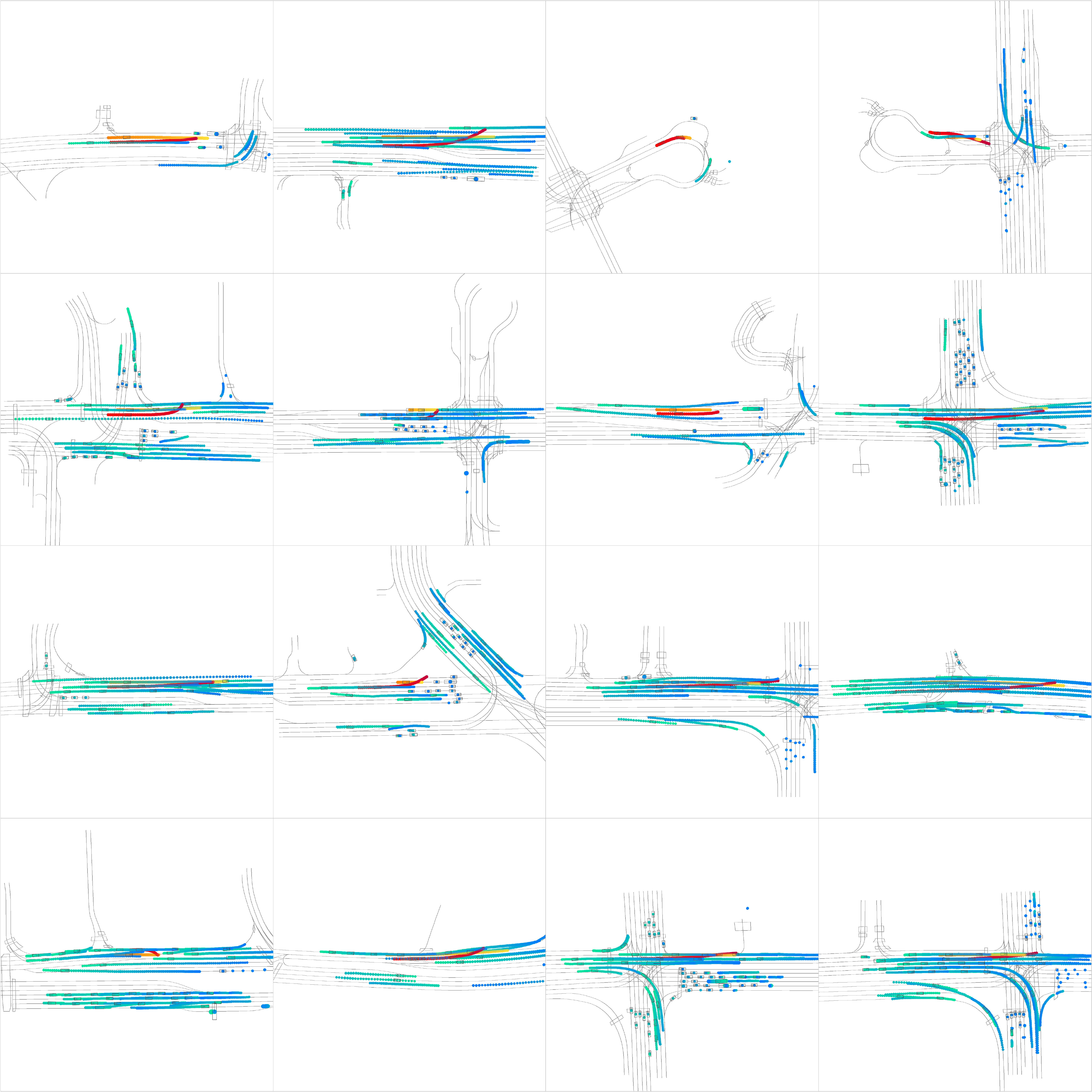}
\caption{\textbf{Various safety-critical layouts from Nexus-Data.}
All scenarios are initialized by the nuPlan~\cite{caesar2021nuplan} and generated by adversarial interactions~\cite{zhang2023cat} within simulators.}
\label{fig:supp-nexus-data}
\end{figure}

\subsection{Layout Diversity Highlights}

We applied handcrafted rules to analyze behavior distributions in nuScenes~\cite{caesar2019nuscenes}, nuPlan~\cite{caesar2021nuplan}, and our Nexus-Data shown in \cref{tab:data_eval}.
For brevity, we omit the proportion of normal forward driving.
Beyond forwarding, turning, and stopping, our dataset demonstrates greater diversity in lane-changing scenarios.
%
%
\cref{fig:supp-nexus-data} visually displays the top-down views of various dangerous driving scenarios, including collisions, quick stops, and reckless merging. 

\subsection{License and Privacy Considerations}

\setcounter{footnote}{0}
All the data is under the CC BY-NC-SA 4.0 license\footnote{https://creativecommons.org/licenses/by-nc-sa/4.0/deed.en}.
Other datasets (including nuPlan \cite{qian2023nuscenes}, Waymo Open \cite{sun2020scalability}, Metadrive~\cite{li2022metadrive}) inherit their own distribution licenses.
We only distribute lane geometries and vehicle trajectories, ensuring compliance with dataset licenses and removing personally identifiable information to prevent privacy risks.

\section{Implementation Details of Nexus}
\label{sec:supp-implementation}

\subsection{Model Design}

As shown in \cref{tab:nexus_arch}, the Nexus architecture is built upon SimpleDiffusion~\cite{hoogeboom2023simple}. 
The model uses rotary embedding for position encoding, which is based on both physical time and denoising steps simultaneously.
The backbone incorporates four layers of TemporalBlock and SpatialBlock, enabling the model to capture temporal and spatial dependencies through attention and feedforward layers.
The Global Encoder uses Perceiver IO~\cite{jaegle2021perceiver} for map feature extraction. 
The final output is projected through a linear layer.

\begin{table}[t]
\centering
\caption{\textbf{Architecture of the Nexus Model.}}
\begin{tabular}{ll}
\toprule
\textbf{Component} & \textbf{Details} \\
\midrule
\textbf{Top-Level Model} & LightningModuleWrapper \\
\midrule
\textbf{Main Model} & Nexus \\
Diffusion Backbone & SimpleDiffusion \\
Cross Attention & LayerNorm (256) + MultiHeadAttention \\
Attention Projection & Linear(256, 256) \\
Input Projection & Linear(25, 256) \\
Timestep Embedder & Linear(256, 256) + SiLU + Linear(256, 256) \\
\midrule
\textbf{Backbone Structure} & CombinedAttention with TemporalBlock \& SpatialBlock \\
TemporalBlock & LayerNorm(256) $\to$ MultiHeadAttention(256) \\
             & $\to$ LayerNorm(256) $\to$ FeedForward MLP \\
             & $\to$ SiLU + Linear(256, 1536) (AdaLN Modulation) \\
SpatialBlock & LayerNorm(256) $\to$ MultiHeadAttention(256) \\
             & $\to$ LayerNorm(256) $\to$ FeedForward MLP \\
             & $\to$ SiLU + Linear(256, 1536) (AdaLN Modulation) \\
MultiHead Attention & Linear(256, 256) with Dropout(0.0) \\
FeedForward MLP & Linear(256, 1024) + GELU + Linear(1024, 256) \\
Final Normalization & LayerNorm(256) \\
Output Projection & Linear(256, 8) \\
\midrule
\textbf{Global Encoder} & PercieverEncoder \\
Cross Attention & LayerNorm(7) + MultiHeadAttention(256) \\
Self-Attention & 2x SelfAttention Blocks \\
\midrule
\textbf{Other Modules} & MapRender, NaivePlanner \\
\bottomrule
\end{tabular}
\label{tab:nexus_arch}
\end{table}

\subsection{Training Details}

Nexus is trained over 1200 hours of real-world driving logs from the nuPlan dataset~\cite{caesar2021nuplan} and 480 hours of collected data from the simulator~\cite{li2022metadrive}.
The training data consists of 10-second driving logs sampled at 2Hz, resulting in 21 frames per sequence (4 historical, 1 current, and 16 future frames). 
The dataset includes 528K scenarios, each covering a 104-meter range.
Training on Waymo~\cite{sun2020scalability} used 531K scenes, each lasting 9 seconds, sampled at 2Hz, with 2 historical frames, 1 current frame, and 16 future frames.
The training task follows a denoising diffusion process, where random noise is added to each agent token across the entire sequence. 
The model is then trained to recover the original sequence, learning to reconstruct motion trajectories under noisy conditions.

We train the model for 80K iterations on 8 GPUs with a batch size of 1024 with AdamW~\cite{loshchilov2018fixing}.
The initial learning rate is $1 \times 10^{-3}$.
We use a learning rate scheduler with a warm-up and cosine decay strategy. 
After the warm-up, the learning rate will gradually decrease according to a cosine function.
The default GPUs in most of our experiments are NVIDIA Tesla A100 devices unless otherwise specified.


\subsection{Sampling Details}
\label{sec:supp-sampling details}

\begin{figure}[t]
    \centering
    \includegraphics[width=\linewidth]{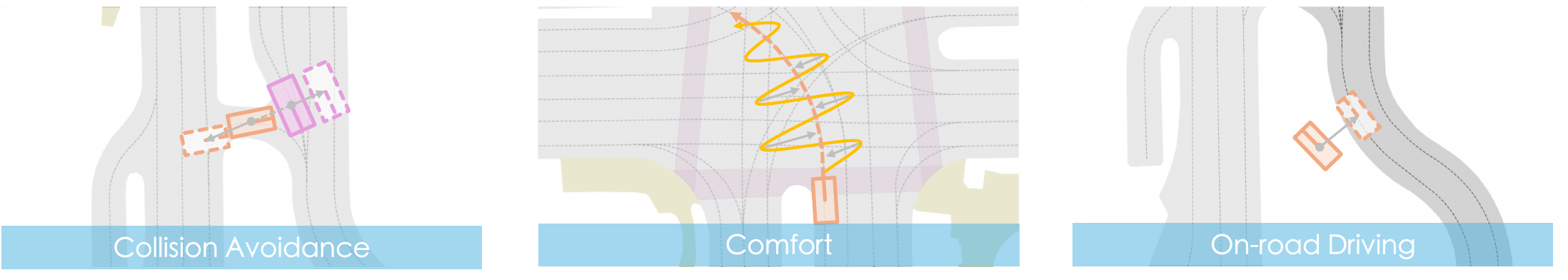}
    \caption{\textbf{Classifier guidance with human-behavior alignment.}
    The three different constraints are applied at each sampling step, contributing to the realism of the generated scenario.}
    \label{fig:constraint}
\end{figure}

\noindent \textbf{Classifier guidance for human-behavior alignment.}
Diffusion models can generate unrealistic driving scenarios due to randomness, requiring human-guided constraints to enhance scene quality.
As shown in \cref{fig:constraint}, we consider three human-behavior rubrics:
1) Collision avoidance: At each step $t$, if two vehicles' bounding boxes overlap, they are pushed apart along their center-connecting line.
It can be written as the following equation:

\begin{align}
f_{\text{collision}}(\mathbf{x}^t, t) &= \big[ \mathbf{x}^{t}_{\text{loc}}, \mathbf{x}^{t, 3:d} \big],\\
\text{where}\ 
\mathbf{x}^{t}_{\text{loc}} &\gets \mathbf{x}^{t}_{\text{loc}} + \lambda_t \sum_{i \neq j} \mathbb{I} \{B(\mathbf{x}^t_i) \cap B(\mathbf{x}^t_j) \neq \varnothing \} \cdot 
\frac{\mathbf{x}^{t}_{i,\text{loc}} - \mathbf{x}^{t}_{\text{j,loc}}}{\|\mathbf{x}^{t}_{i,\text{loc}} - \mathbf{x}^{t}_{j,\text{loc}}\|},
\end{align}
where $\lambda_t$ is a scalar coefficient used to control the extent of separation at time $t$.
$\mathbb{I}$ is an indicator function that takes the value 1 when the bounding boxes of vehicle $i$ and vehicle $j$ overlap and 0 otherwise.
$B$ is the function used to form the vehicle's bounding box.
The fractional term represents the unit direction vector of the centerline between vehicle $i$ and vehicle $j$.

2) Comfort: Enforcing smooth longitudinal and lateral accelerations by averaging adjacent trajectory points.

\begin{align}
f_{\text{comfort}}(\mathbf{x}^t, t) &= \big[ \mathbf{x}^{t}_{\text{loc}}, \mathbf{x}^{t, 3:d} \big],\\
\text{where}\ 
\mathbf{x}^{t}_{\text{loc}} &\gets \mathbf{x}^{t}_{\text{loc}} - \lambda_t \mathbf{a}^{t}, \\
\mathbf{a}^{t} &= \frac{1}{2} (\mathbf{x}^{t}_{\tau-1,\text{loc}} - 2\mathbf{x}^{t}_{\tau, \text{loc}} + \mathbf{x}^{t}_{\tau+1, \text{loc}}).
\end{align}

First, the longitudinal and lateral accelerations $a^{t}$ are approximated using the second-order difference at time $\tau$ and smoothed by averaging adjacent trajectory points.  
Then, the trajectory is refined by subtracting a proportion $lambda_t$ of the acceleration, reducing abrupt speed changes for smoother motion.

3) On-road driving: Pull the vehicle toward the nearest centerline point when it strays too far.

\begin{align}
f_{\text{on road}}(\mathbf{x}^t, t) &= \big[ \mathbf{x}^{t}_{\text{loc}}, \mathbf{x}^{t, 3:d} \big],\\
\text{where}\
\mathbf{x}_{i,\text{loc}}^{t} & \gets \mathbf{x}_{i,\text{loc}}^{t} + \lambda_t \mathbb{I} \{\|\mathbf{x}_{i,\text{loc}}^{t} - \mathbf{c}_i^t\| > d_{\text{th}}\} \cdot (\mathbf{c}_i^t - \mathbf{x}_{i, \text{loc}}^{t}), \\
\mathbf{c}_i^t &= \operatorname{argmin}_{l, n} \|\mathbf{x}_{i, \text{loc}}^{t} - \mathbf{c}_{l, n, \text{loc}}\|.
\end{align}

The vehicle identifies the closest lane point $\mathbf{c}_{i}^t$ among all points $\mathbf{c} \in \mathbb{R}^{L \times N \times D}$ by minimizing the Euclidean distance using $ \arg\min_{l,n}$.  
When the deviation exceeds the threshold $ d_{\text{th}} $, the vehicle adjusts its position by moving from $ \mathbf{x}^t_{i,\text{loc}} $ toward the closest centerline point $ \mathbf{c}_i^t $, with the adjustment magnitude controlled by $\lambda_t$.

\noindent \textbf{Sampling.}
The sampling process is inherited from SimpleDiffusion~\cite{hoogeboom2023simple}.
It starts with random Gaussian noise and is performed by Denoising Diffusion Implicit Models (DDIM)~\cite{song2020DDIM} for 32 steps.
For classifier guidance~\cite{saharia2022photorealistic}, we set the total value of $\lambda$ for the three constraints to be 0.2. 
If more than two constraints are active simultaneously, the value of $\lambda$ will be evenly distributed among them.
The sampling speed is \textbf{206} milliseconds per step per batch.

\section{Experiments}
\label{sec:supp-experiments}

We conduct extensive experiments on multiple datasets to evaluate the performance of our method.
Our baseline is built on a reproduced full-sequence training Diffusion Policy~\cite{chi2023diffusion}.
For comparison convenience, we trained two models on the nuPlan and Nexus-Data datasets, respectively, namely Nexus-\textit{Full} and Nexus, adopting the same training strategy.

\subsection{Protocols and Metrics}

\noindent\textbf{ADE:} 
It measures the average displacement differences between the generated and ground truth trajectories, excluding goal points and invalid trajectory points from the calculation.

\noindent\textbf{$R_{\text{road}}$:}
It measures the off-road rate of vehicles in the generated scenes. Off-road instances are detected by checking whether a vehicle's center deviates from its assigned centerline at each timestep. The rate is calculated as the number of vehicles that have gone off-road divided by the total number of valid vehicles.

\noindent\textbf{$R_{\text{col}}$:}
It measures the collision rate among agents in the generated scenes. Collisions are detected by checking for overlaps between agent bounding boxes at each timestep. The rate is calculated as the number of collided vehicles divided by the total number of valid vehicles.

\noindent\textbf{$M_{\text{k}}$:}
It measures the stability of generated trajectories using the average of four metrics: tangential and normal acceleration along the heading at each timestep and their derivatives (jerk). 
Lower values indicate smoother and more comfortable trajectories.

\noindent\textbf{Composite Metric:}
It is a comprehensive metric for evaluating the realism of scene generation. During evaluation, the model generates a scene 32 times based on a 1-second history, forming a distribution. The likelihood between this distribution and the ground truth is then computed across factors like speed, distance, and collisions.

\noindent\textbf{Score, $S_{\text{col}}$, and $S_{\text{p}}$:}
It is the main metric for assessing the reasonableness of agent planning trajectories in the nuPlan closed-loop evaluation.
The metric considers comfort, collisions, road adherence, lane changes, and mileage completion, scoring 1 for success and 0 for failure per scenario.
$S_{\text{col}}$ and $S_{\text{p}}$ are sub-metrics measuring trajectory collision rate and distance completion rate, respectively.

\subsection{Evaluation Tasks}

\noindent \textbf{Free exploration.}
This task conditions the past 2 seconds of all vehicle states and lane centerlines from the nuPlan driving log to freely generate an 8-second future scene at a 0.5-second time interval.
In the generation process, the noise level of each token is determined by the scheduling strategy.
Invalid vehicles at the corresponding timestep are ignored.
In the experiments, we used off-the-shelf IDM~\cite{treiber2000congested} and GUMP~\cite{hu2025solving}, as well as our implementations of Diffusion Policy~\cite{chi2023diffusion} and SceneDiffuser~\cite{jiang2024scenediffuser}.

\noindent \textbf{Conditioned generation.}
On top of free exploration, goal points are added to valid vehicles by setting the token's noise level at that timestep to 0 during inference.

\noindent \textbf{Waymo open sim agent evaluation.}
The Waymo evaluation requires generating 32 future scene predictions based on 1 second of historical observations, including vehicles, pedestrians, and cyclists. The evaluation is conducted at 10Hz, and we interpolate the 2Hz model to match the required scene frequency.

\noindent \textbf{Closed-loop evaluation.}
In the nuPlan closed-loop evaluation, the environment and agent are treated separately.
The agent predicts an 8-second trajectory based on 2 seconds of historical observations and takes 0.1-second actions. The environment updates the scene based on the agent's actions, running at 10Hz.
In the experiment using the generative model as a world generator, we replace the original nuPlan environment. 
Starting with a 2-second historical scene, it generates and updates the next scene (0.1 seconds ahead) based on the agent's actions.
In the experiment using synthetic data to augment the planner, we train the agent with different amounts of synthetic and real data and then evaluate it in the nuPlan closed-loop environment.

\subsection{Generation of Novel Scenarios}
\label{sec:novel scenario}

Nexus can serve as a data engine to automatically generate new scenarios in batches. 
Specifically, we use the first two seconds of nuPlan raw logs as initial conditions and generate new scenarios through free exploration, conditioned generation, and attacks on the ego vehicle.  
For attack-based scenario generation, we follow a similar approach to \cref{sec:Nexus-Data} to select attacking vehicles. 
For goal point selection, we define a sector along the historical trajectory direction of a chosen attack vehicle, with the sector's radius determined by speed and an angle $\alpha$. 
The future positions of other vehicles within this sector represent highly probable goal points that could lead to a collision.  
During generation, we maintain a 4:4:2 ratio among the three types of scenario data to ensure a balanced distribution of scenarios.



\begin{figure*}[]
\begin{center}
\centerline{\includegraphics[width=\linewidth]{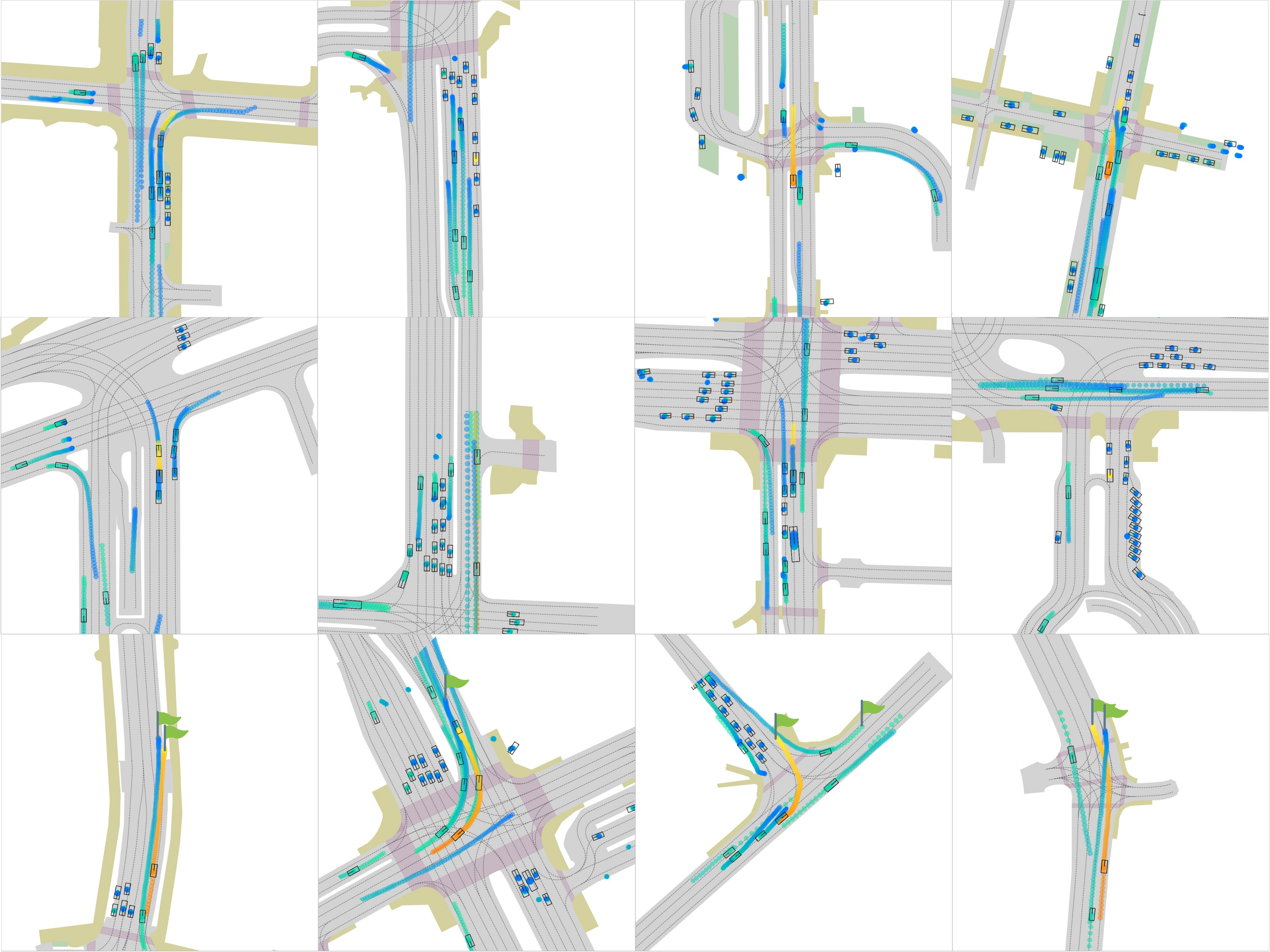}}
\caption{
\textbf{More visualizations of Nexus on free exploration and conditioned generation.}
}
\label{fig:supp-main-vis}
\end{center}
\end{figure*}

\begin{figure*}[]
\begin{center}
\centerline{\includegraphics[width=\linewidth]{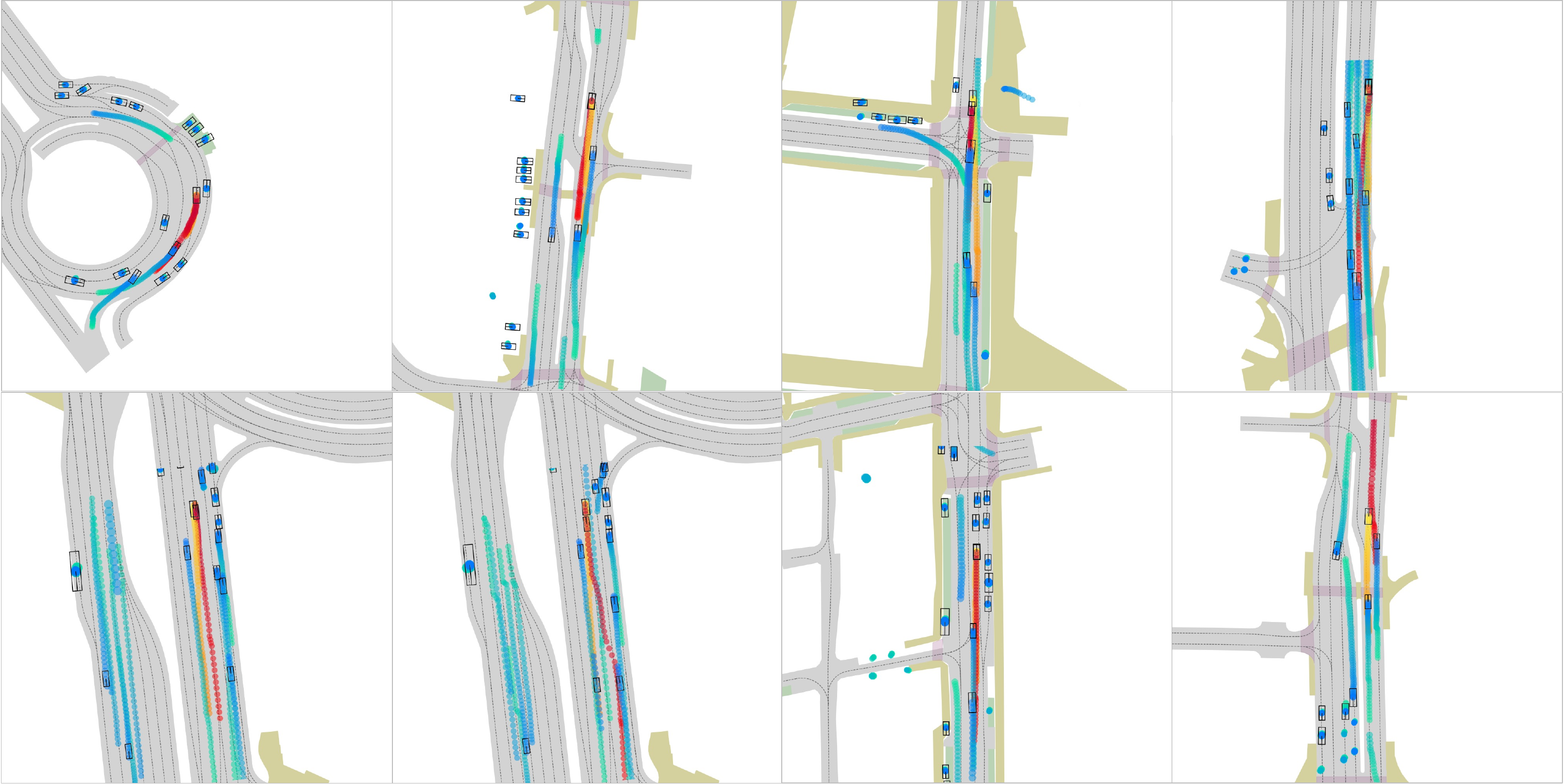}}
\caption{
\textbf{Applications of Nexus for generating diverse corner cases in autonomous driving.}
}
\label{fig:supp-attack-vis}
\end{center}
\end{figure*}

\begin{figure*}[]
\begin{center}
\centerline{\includegraphics[width=\linewidth]{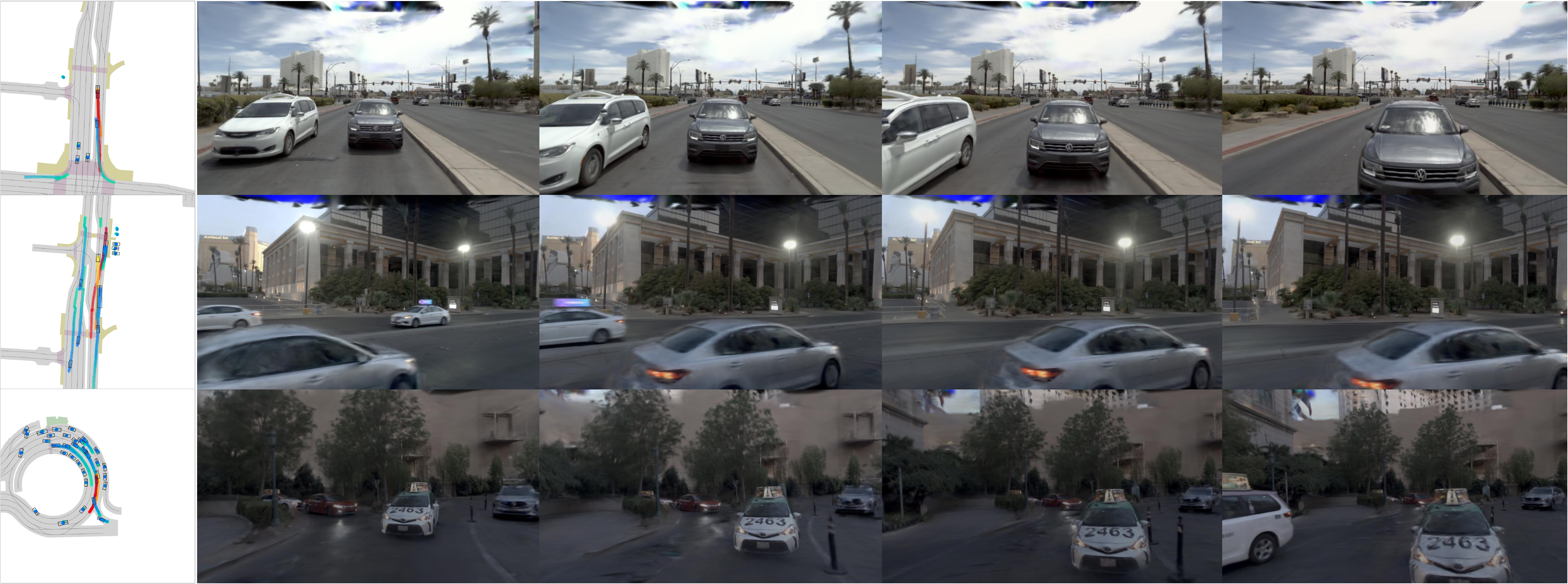}}
\caption{
\textbf{Leveraging neural radiance fields to provide realistic visual appearances for scenes generated by Nexus.}
On the left is the bird's-eye-view layout, and on the right is the rendered scene.
}
\label{fig:supp-nerf-vis}
\end{center}
\end{figure*}

\subsection{Qualitive Results}
Thanks to the decoupled diffusion structure, Nexus can transition among free exploration, conditioned generation, and diverse corner case synthesis seamlessly, enabling adaptive scene generation. 
Moreover, leveraging the neural rendering field (NeRF)~\cite{tonderski2024neurad}, Nexus transforms generated traffic layouts into photorealistic scenes, enabling controllable visual synthesis.

\noindent \textbf{Free exploration and conditioned generation.}
\cref{fig:supp-main-vis} showcases Nexus's versatility in generating driving scenarios. The first two rows depict free exploration, where decoupled noise states enable diverse traffic layouts without explicit conditioning, capturing complex interactions and behaviors. The last row illustrates conditioned generation, where low-noise target tokens guide scene evolution toward goal states (green flags), enhancing controllability and reactivity while maintaining realism.

\noindent \textbf{Diverse corner case generation.}
As shown in \cref{fig:supp-attack-vis}, Nexus generates diverse corner cases, including abrupt cut-ins, sudden braking, and potential collisions. This strengthens Nexus's utility for training robust autonomous systems.

\noindent \textbf{Controllable visual rendering with NeRF integration.}
\cref{fig:supp-nerf-vis} showcases NeRF-based rendering, converting Nexus-generated layouts into photorealistic scenes. The left panel depicts the bird's-eye view, while the right presents the rendered scene, demonstrating controllable visual synthesis for interactive simulations and closed-loop evaluation.

\subsection{Failure Cases}
\begin{figure*}[]
\begin{center}
\centerline{\includegraphics[width=\linewidth]{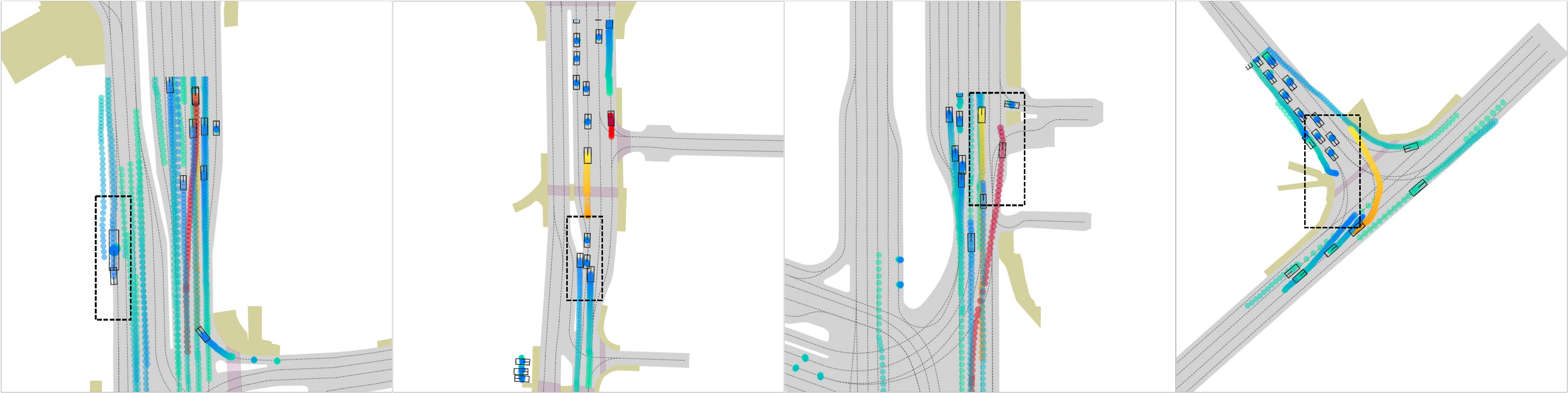}}
\caption{
\textbf{Failure cases of Nexus.}
}
\label{fig:supp-fail-case}
\end{center}
\end{figure*}

\cref{fig:supp-fail-case} illustrates Nexus failure cases. The two left cases show incorrect collisions: one between a bus and a sedan and another with overlapping bounding boxes. The two right cases highlight the model's difficulty in making decisions in complex road networks due to limited map information. Future improvements will include adaptive collision awareness and the addition of road boundaries and drivable areas to address these issues.

\end{document}